\documentclass[pdflatex,sn-nature]{sn-jnl}

\usepackage{silence}
\usepackage{graphicx}
\usepackage{booktabs}
\usepackage{multirow}
\usepackage{float}
 \usepackage{subcaption}

\usepackage{amsmath}
\usepackage{amssymb}
\usepackage{amsfonts}
\usepackage{amsthm}
\usepackage{mathrsfs}

\usepackage{algorithm}
\usepackage{algorithmicx}
\usepackage{algpseudocode}

\usepackage{xcolor}
\usepackage{textcomp}
\usepackage{manyfoot}
\usepackage{listings}

\hypersetup{
  hidelinks,
  bookmarksdepth=3
}

\DeclareFontFamily{U}{rsfs}{}
\DeclareFontShape{U}{rsfs}{m}{n}{<->rsfs10}{}

\theoremstyle{thmstyleone}%
\theoremstyle{thmstyletwo}%

\theoremstyle{thmstylethree}%

\begin{document}

\title[FADA: Knowledge-Distilled VLMs for Fetal Ultrasound]{FADA: Accessible fetal ultrasound interpretation and annotation with a selectively distilled unified vision-language model}


\author*[1]{\fnm{Mahmood} \sur{Alzubaidi}}\email{malzubaidi@hbku.edu.qa}
\author[1 ]{\fnm{Uzair} \sur{Shah}}
\author[1 ]{\fnm{Raden} \sur{Muaz}}
\author[1]{\fnm{Ines} \sur{Abbes}}
\author[2]{\fnm{Nader} \sur{Mohammed}}
\author[3]{\fnm{Abdullatif} \sur{Magram}}
\author[4]{\fnm{Khalid} \sur{Alyafei}}
\author[1]{\fnm{Mowafa} \sur{Househ}}
\author*[1]{\fnm{Marco} \sur{Agus}}

\affil[1]{\orgdiv{College of Science and Engineering }, \orgname{Hamad Bin Khalifa University}, \orgaddress{\city{Doha}, \country{Qatar}}}
\affil[2]{\orgdiv{Center for Clinical Precision Medicine and Genomics}, \orgname{HMC}, \orgaddress{\city{Doha}, \country{Qatar}}}
\affil[3]{\orgdiv{Advanced AlRazi Diagnostic Center}, \orgaddress{\city{Al-Hodeidah}, \country{Yemen}}}
\affil[4]{\orgdiv{Sidra Medicine}, \orgaddress{\city{Doha}, \country{Qatar}}}

\abstract{
A global shortage of trained sonographers limits prenatal ultrasound screening in low- and middle-income countries, where over half of pregnant women receive no skilled sonography. Current deep learning approaches address detection, segmentation, or classification in isolation, each demanding a separate model and expert-specified labels at inference. We present FADA, a unified vision-language model built on Qwen3.5-VL that performs clinical interpretation, classification, detection, and segmentation through a single interpretation-first pipeline without external labels. FADA distills knowledge from four domain-specific foundation models (FetalCLIP, UltraSAM, USF-MAE, UltraFedFM) via offline pre-computed feature caching. Selective distillation, which applies feature alignment only to annotation tasks while interpretation relies on standard fine-tuning, consistently outperforms full distillation across most evaluation axes. The recommended variant, FADA-SKD, achieves 0.8820 mean Dice for segmentation, 0.7671 mAP@0.50 for detection, and 100\% structured interpretation compliance. Expert sonographer validation across 237 images confirms clinically acceptable outputs in both autonomous and human-in-the-loop modes, with 73.5\% of interpretations scoring perfectly under clinician guidance. The system is trainable on a single consumer GPU and deployable without cloud connectivity. We validate edge deployment by running the compressed 0.8B model on a commodity smartphone (Qualcomm Snapdragon~7 Gen~1, 12~GB RAM) using llama.cpp with GGUF quantization, completing the full 5-phase pipeline in approximately 60 seconds entirely offline. This establishes a practical pathway for integrating AI-assisted fetal assessment with portable ultrasound devices in a stand-alone fashion, directly addressing diagnostic access gaps in resource-constrained settings. Code, models, and data are available at \url{https://github.com/mahmoodphd/FADA}
}

\keywords{Fetal ultrasound, Vision-language model , Knowledge distillation , Low-resource settings , Medical image interpretation}

\maketitle


\section{Introduction}\label{sec:intro}

Fetal ultrasound remains the cornerstone of prenatal anatomical assessment worldwide, yet the World Health Organization estimates that over half of pregnant women in low- and middle-income countries (LMICs) receive no skilled sonography during pregnancy~\cite{who2016recommendations}. This disparity arises primarily from a critical shortage of trained sonographers: some sub-Saharan African countries report fewer than one sonographer per 100,000 population, compounded by high equipment maintenance costs and limited diagnostic support infrastructure in rural facilities~\cite{kim2017obstetric}. The resulting gap in prenatal screening disproportionately contributes to preventable perinatal morbidity and mortality in the regions bearing the highest burden of adverse obstetric outcomes. Bridging this gap demands AI solutions that are not merely accurate but explicitly designed for deployment without specialist infrastructure: trainable on consumer hardware, deployable without cloud connectivity, and operable by non-specialist health workers with remote expert oversight.

Deep learning methods now achieve strong performance on individual fetal ultrasound tasks, including plane classification~\cite{burgos2020evaluation}, anatomical structure detection~\cite{chen2023anatomical}, and biometric measurement segmentation~\cite{van2018automated}. These approaches are constrained by their task-specific design: each requires a separate model, task-specific training data, and expert-curated class labels at inference to specify target structures. Such a paradigm is ill-suited for settings where the very expertise needed to guide these models is the resource that is scarce.

Vision-language models (VLMs) offer an alternative by unifying multiple vision tasks within a single model prompted with natural language. Recent work demonstrates VLM potential for medical imaging~\cite{li2024llava}, including specialized models for ultrasound~\cite{jin2026ultrasoundclip} and fetal imaging~\cite{he2025fetalmind}. Yet no existing system provides a unified pipeline that autonomously interprets a fetal ultrasound image, identifies appropriate anatomical structures for analysis, and performs targeted detection and segmentation without requiring external class labels.

Here we present FADA (Fetal Anatomy Delineation and Analysis), a unified VLM that addresses these limitations through an interpretation-first architecture. Given a fetal ultrasound image, FADA autonomously generates a structured clinical interpretation, determines appropriate anatomical targets, and performs detection and segmentation within a single forward pass, without requiring operator-specified class labels. The principal contributions of this work are:

\begin{enumerate}
    \item \textbf{Unified multi-task architecture.} FADA performs clinical interpretation, anatomical classification, bounding-box detection, and polygon segmentation within a single model through a 5-phase pipeline. By generating the clinical interpretation first, the model determines which structures to detect and segment based on its own assessment of the anatomy, eliminating the need for external class labels at inference.
    
    \item \textbf{Selective knowledge distillation.} Feature-level alignment from four domain-specific teachers (FetalCLIP, UltraSAM, USF-MAE, UltraFedFM) is applied exclusively to annotation training data while interpretation training receives only supervised fine-tuning. This selective strategy outperforms full distillation across segmentation, detection, classification, and expert-rated interpretation quality, indicating that spatial teacher features and language generation benefit from distinct training regimes.
    
    \item \textbf{Offline distillation with pre-computed caching.} Teacher features are extracted once and stored in HDF5 format (453K vectors across 4 teachers $\times$ 3 layers), eliminating concurrent teacher inference and reducing GPU memory requirements by approximately 60\%. This enables knowledge distillation from large foundation models on a single consumer GPU.
    
    \item \textbf{Cross-task knowledge transfer.} FADA produces clinically meaningful interpretations for anatomical categories encountered only during annotation training (e.g., FUSEP brain structures, FOCUS cardiac anatomy), indicating that detection and segmentation supervision transfers interpretive knowledge through the shared visual encoder.
    
    \item \textbf{Expert-validated dual deployment.} Both fully autonomous and human-in-the-loop modes are validated by an expert sonographer across 237 images and 49 clinical cases, achieving 73.5\% perfect interpretation scores under clinician guidance and establishing clinical viability for decision support in resource-constrained settings.
    
    \item \textbf{Validated edge deployment on commodity hardware.} Model weights, training code, web application, interpretation dataset, and a compressed 0.8B mobile variant (GGUF Q4\_K\_M quantization) are released under open licenses. We demonstrate end-to-end on-device inference on a commodity Android smartphone using llama.cpp, completing the full 5-phase pipeline in ${\sim}$60\,s without cloud connectivity, establishing that the model can be integrated with portable fetal ultrasound devices in a stand-alone fashion.
\end{enumerate}


\section{Related Work}\label{sec:related}

\subsection{Vision-Language Models in Medical Imaging}

Vision-language models (VLMs) have reshaped medical image analysis by unifying previously disparate visual recognition tasks within a single generative framework~\cite{ryu2025vision,kalpelbe2025vision}. LLaVA-Med~\cite{li2024llava} showed that a general-purpose VLM could be adapted for biomedical question answering with minimal domain-specific training. More recently, Dolphin~\cite{wang2025dolphin} introduced a multimodal large language model specifically for ultrasound understanding, underscoring the potential of domain-specialized VLMs. These models, however, primarily address single-task scenarios (e.g., report generation or classification) and do not integrate spatial grounding tasks such as detection and segmentation within the same generative pipeline.

\subsection{Knowledge Distillation for Medical AI}

Knowledge distillation (KD) has become a standard technique for deploying high-performance medical imaging models under computational constraints~\cite{li2025knowledge}. Recent work extends single-teacher distillation to multi-teacher frameworks, where complementary expertise from multiple foundation models is transferred to a compact student~\cite{xun2025multipleTeacher}. While these approaches succeed in classification and segmentation tasks independently, their application to multi-task VLMs, where the student must simultaneously learn visual grounding and language generation, remains unexplored. FADA addresses this limitation through selective distillation that conditions teacher alignment on task type, preventing interference between spatial and linguistic learning objectives.

\subsection{Fetal Ultrasound AI}

Deep learning for fetal ultrasound has advanced from single-task models for biometric measurement~\cite{van2023fetal} and anatomical structure detection~\cite{chen2023anatomical} to multi-task systems capable of end-to-end assessment~\cite{benson2025fetal,bai2025beyond}. FetalMind~\cite{he2025fetalmind} represents a closely related effort, employing a large VLM with disease-view bipartite graphs for structured fetal neurosonography reporting. FetalMind requires multiple views and separate diagnostic modules for different assessment aspects, limiting applicability in settings where only single images are available. Ultrasound-CLIP~\cite{jin2026ultrasoundclip} achieves strong ultrasound-text alignment through heterogeneous graph encoding and contrastive learning, but produces embeddings rather than clinically actionable structured outputs.

Concurrently, SonoMate~\cite{guo2025sonomate} introduced a visually grounded language model for fetal ultrasound understanding using video-text alignment in Nature Biomedical Engineering. SonoMate demonstrates strong detection performance but focuses on video-level understanding rather than comprehensive per-image multi-task analysis. FetalCLIP~\cite{fetalclip} provides a dedicated visual-language foundation model for fetal ultrasound (427M parameters), while MobileFetalCLIP~\cite{saeed2025mobilefetalclip} distills FetalCLIP to mobile scale using diagonal-anchored repulsive KD (DARK), achieving 88.6\% HC18 validity. Both address classification only rather than the full pipeline of interpretation, detection, segmentation, and keypoint localization.

In contrast to these systems, FADA provides a single-image, single-model pipeline that autonomously interprets, detects, and segments without requiring external class labels or multiple imaging views, a design driven by the practical constraints of low-resource clinical environments.

\subsection{Clinical AI Validation and Deployment}

Translating AI systems from benchmarks to clinical practice demands rigorous validation beyond automated metrics~\cite{hu2025human}. Human-in-the-loop (HiL) evaluation enables expert oversight while maintaining workflow efficiency, and has become integral to responsible clinical AI deployment~\cite{jmir2026pocus}. Recent studies highlight specific barriers to deploying AI in point-of-care ultrasound, including data heterogeneity, device variability, and the need for robust out-of-distribution generalization~\cite{vega2025barriers}. Task-shifting approaches, where AI enables non-specialist operators to perform screening tasks previously requiring experts, have shown promise for obstetric ultrasound in LMICs~\cite{dellaripa2025obstetric,taskshift2022pocus}. For resource-constrained settings, mobile health (mHealth) platforms leveraging AI-driven edge computing offer practical pathways for healthcare delivery where traditional infrastructure is unavailable~\cite{recent2025advances,edge2025transforming}. Knowledge distillation for mobile VLMs has also progressed substantially, with cross-modal alignment techniques enabling deployment on resource-limited devices~\cite{feng2025alignkd}. FADA is designed with these deployment realities in mind, offering both cloud-based web deployment and a compressed 0.8B model for offline edge inference.


\section{Results}\label{sec:results}

\subsection{Quantitative Performance}\label{sec:quantitative}

Five model variants were evaluated: FADA-Base and FADA-SKD at both 4B and 0.8B parameter scales, plus FADA-FKD at 4B, on a held-out test set of 4,478 samples spanning detection (1,463), segmentation (544), classification (2,400), and keypoint localization (71) tasks across 8 source datasets. Table~\ref{tab:main_results} summarizes the results.

\begin{table}[!t]
\caption{Quantitative comparison of FADA model variants on 4,478 test samples (1,000 bootstrap iterations, 95\% CI). FADA-SKD applies distillation only to annotation data; FADA-FKD applies it to all data including interpretation. Best values per metric within each scale are \textbf{bolded}.}\label{tab:main_results}
\begin{tabular*}{\textwidth}{@{\extracolsep\fill}lccccc}
\toprule
\textbf{Model} & \textbf{mAP@0.50} & \textbf{mAP@0.75} & \textbf{Dice} & \textbf{IoU} & \textbf{Cls Acc} \\
\midrule
FADA-Base (4B)   & \textbf{0.780}$\pm$0.022 & 0.421$\pm$0.024 & 0.881$\pm$0.028 & 0.813$\pm$0.032 & 0.823$\pm$0.015 \\
FADA-SKD (4B)    & 0.767$\pm$0.022 & 0.440$\pm$0.026 & \textbf{0.882}$\pm$0.026 & \textbf{0.815}$\pm$0.031 & \textbf{0.838}$\pm$0.015 \\
FADA-FKD (4B)    & 0.770$\pm$0.022 & \textbf{0.458}$\pm$0.026 & 0.879$\pm$0.026 & 0.811$\pm$0.033 & 0.830$\pm$0.015 \\
\midrule
FADA-Base (0.8B) & \textbf{0.689}$\pm$0.024 & \textbf{0.382}$\pm$0.025 & 0.863$\pm$0.030 & 0.790$\pm$0.034 & 0.838$\pm$0.016 \\
FADA-SKD (0.8B)  & 0.674$\pm$0.023 & 0.376$\pm$0.026 & \textbf{0.866}$\pm$0.028 & \textbf{0.792}$\pm$0.032 & \textbf{0.843}$\pm$0.015 \\
\bottomrule
\end{tabular*}
\end{table}

FADA-SKD (4B) achieves the best overall segmentation performance (Dice: 0.8820, 95\% CI: $\pm$0.028; IoU: 0.8149, 95\% CI: $\pm$0.032) and the highest classification accuracy among distilled variants (0.8379), while all model variants maintain 100\% structured JSON interpretation compliance. Statistical analysis confirms that SKD preserves detection performance comparable to the Base model ($\Delta$mAP@0.50 = $-$0.013, $p$=0.41). FADA-FKD achieves the best mAP@0.50 (0.7695) and mAP@0.75 (0.4576) among KD variants, but at the cost of lower classification accuracy (0.8296) and marginally reduced segmentation (Dice: 0.8790). The detection difference between FKD and Base is minor ($\Delta$mAP@0.50 = $-$0.010). This pattern validates the selective distillation strategy: spatial teacher features benefit detection at fine-grained thresholds while selective application preserves segmentation quality and classification accuracy.

\begin{figure}[!t]
    \centering
    \includegraphics[width=\textwidth]{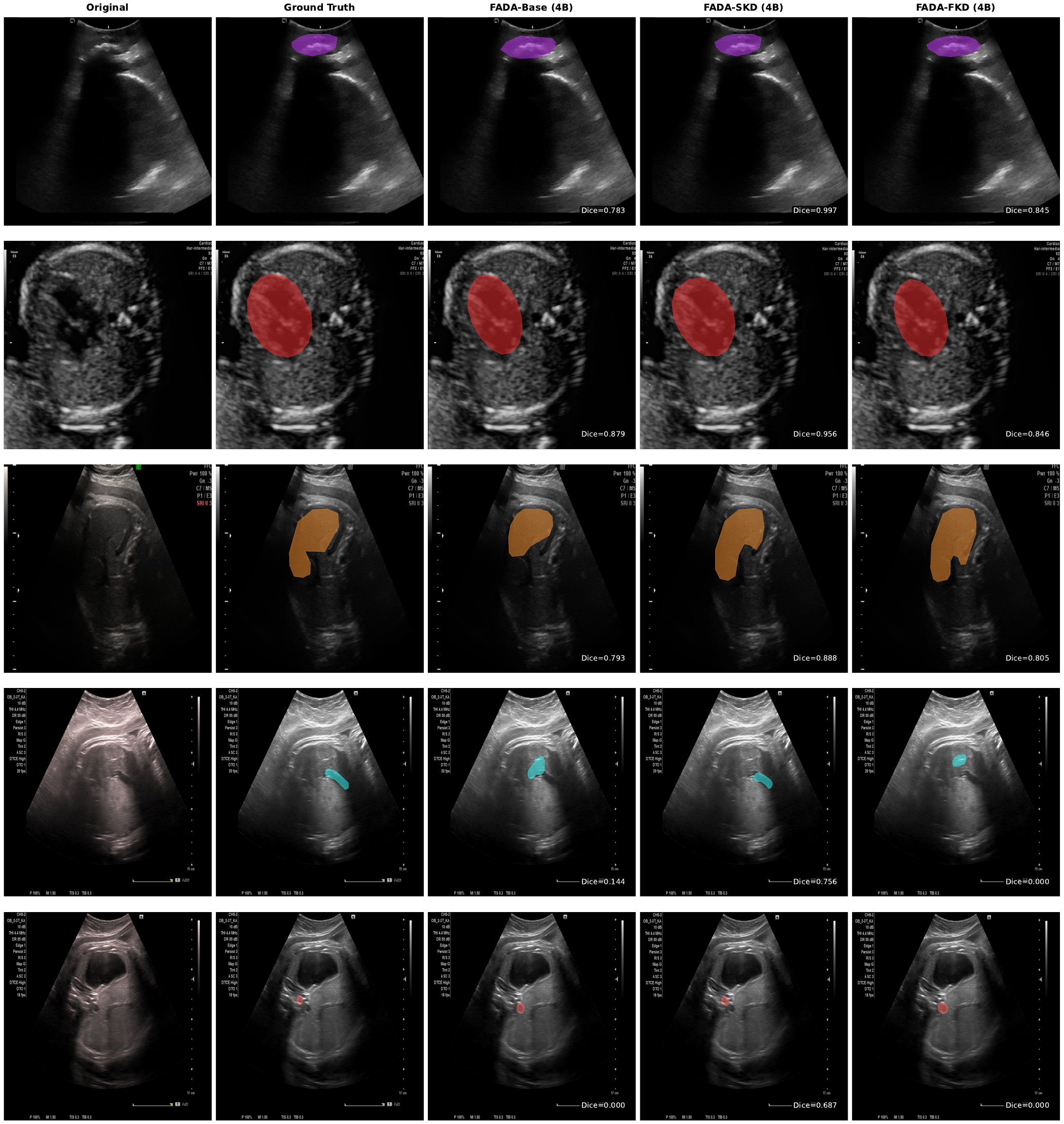}
    \caption{Segmentation ground truth vs prediction comparison across FADA model variants. Each row shows a different anatomical structure (pubic symphysis, cardiac, liver, stomach, artery); columns display the original ultrasound image, ground truth segmentation mask, and predictions from FADA-Base, FADA-SKD, and FADA-FKD (all 4B). Dice coefficients are annotated per prediction panel. Images were selected via quantitative analysis to highlight cases where FADA-SKD achieves the largest advantage: FADA-SKD detects structures entirely missed by other variants (artery: +0.687, stomach: +0.613 Dice advantage) and produces substantially tighter boundaries for pubic symphysis (+0.152) and liver (+0.084).}
    \label{fig:seg_comparison}
\end{figure}

For the 0.8B variants intended for edge deployment, selective KD yields consistent improvements in segmentation and classification: Dice rises from 0.8625 to 0.8662 and classification accuracy from 0.8375 to 0.8433, confirming that SKD generalizes across model scales. Detection mAP@0.50 decreases slightly (0.6885 to 0.6744), consistent with the 4B pattern where feature distillation trades coarse localization accuracy for finer boundary precision. Figure~\ref{fig:seg_comparison} provides visual comparison of segmentation predictions against ground truth, illustrating FADA-SKD's superior boundary adherence.

\paragraph{Cross-Task Consistency.}
Detection-segmentation correlation analysis across model variants reveals that FADA-FKD achieves the highest cross-task consistency (Pearson $r$=0.74) followed by FADA-SKD ($r$=0.61) and FADA-Base ($r$=0.55). While FKD exhibits the tightest detection-segmentation coupling, SKD achieves the best absolute segmentation performance and expert-rated interpretation quality, making it the preferred deployment variant. All segmentation differences between model variants are not statistically significant ($p>$0.90), confirming that the primary effect of distillation strategy falls on the detection-interpretation trade-off rather than spatial precision. A detailed failure mode analysis (Supplementary Table~S4) reveals that interpretation failures concentrate in out-of-distribution anatomy (aorta views, pubic symphysis) and correlate strongly with dataset-level coverage during training.

\begin{figure}[!t]
    \centering
    \includegraphics[width=\textwidth]{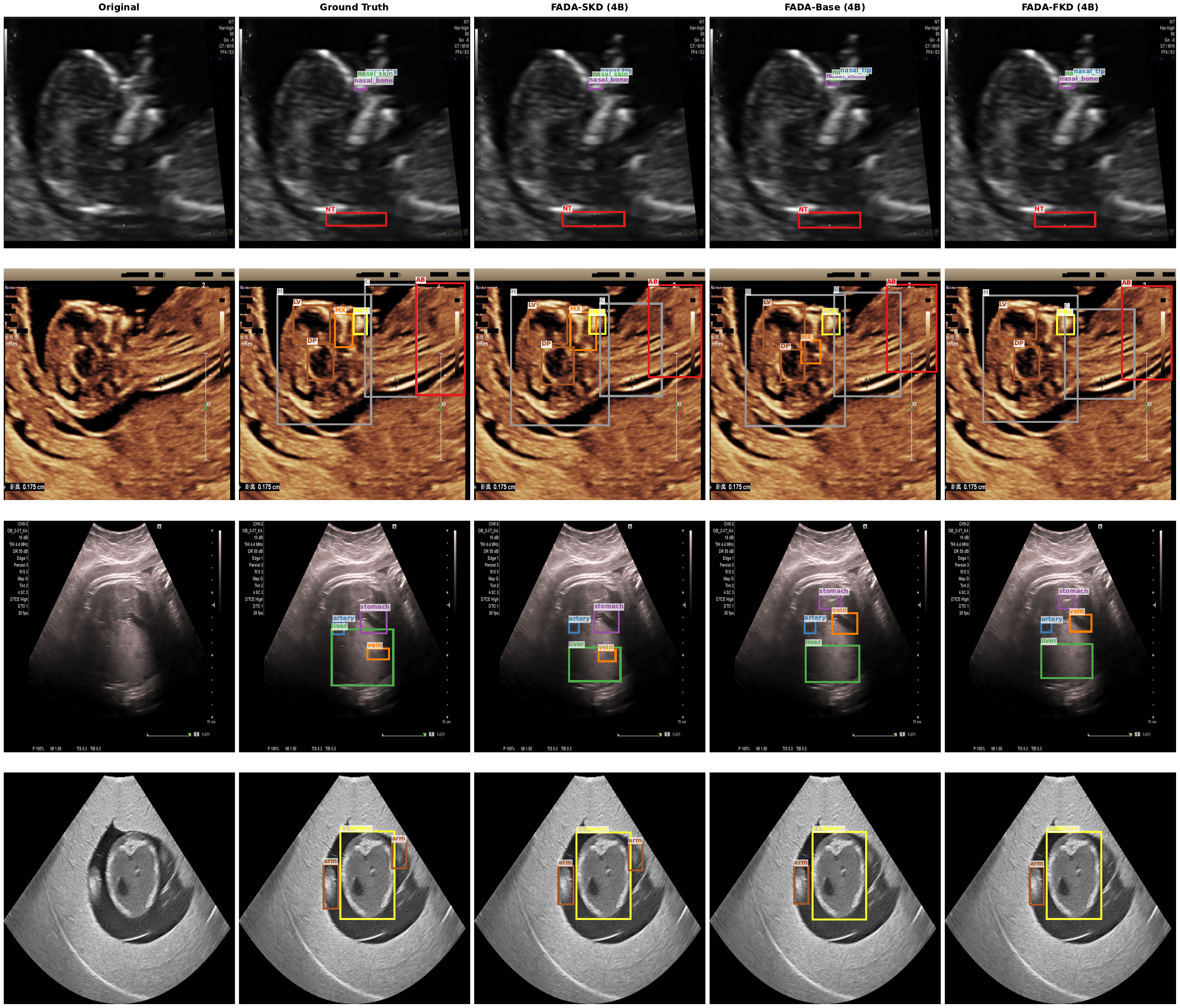}
    \caption{Detection ground truth vs prediction comparison. Each row shows a representative case from a different source dataset (CRL/NT, FUSEP brain, fetal abdominal, FPUS23) with multiple anatomical structures. Columns: original ultrasound image, ground truth bounding boxes, FADA-SKD (4B) predictions, FADA-Base (4B) predictions, and FADA-FKD (4B) predictions. Bounding boxes are color-coded by structure class. Images were selected via quantitative analysis: FADA-SKD detects 4/4 structures where Base and FKD detect only 1/4 (row~1), and achieves higher mean IoU across all selected cases.}
    \label{fig:det_comparison}
\end{figure}

\begin{figure}[!t]
    \centering
    \includegraphics[width=\textwidth]{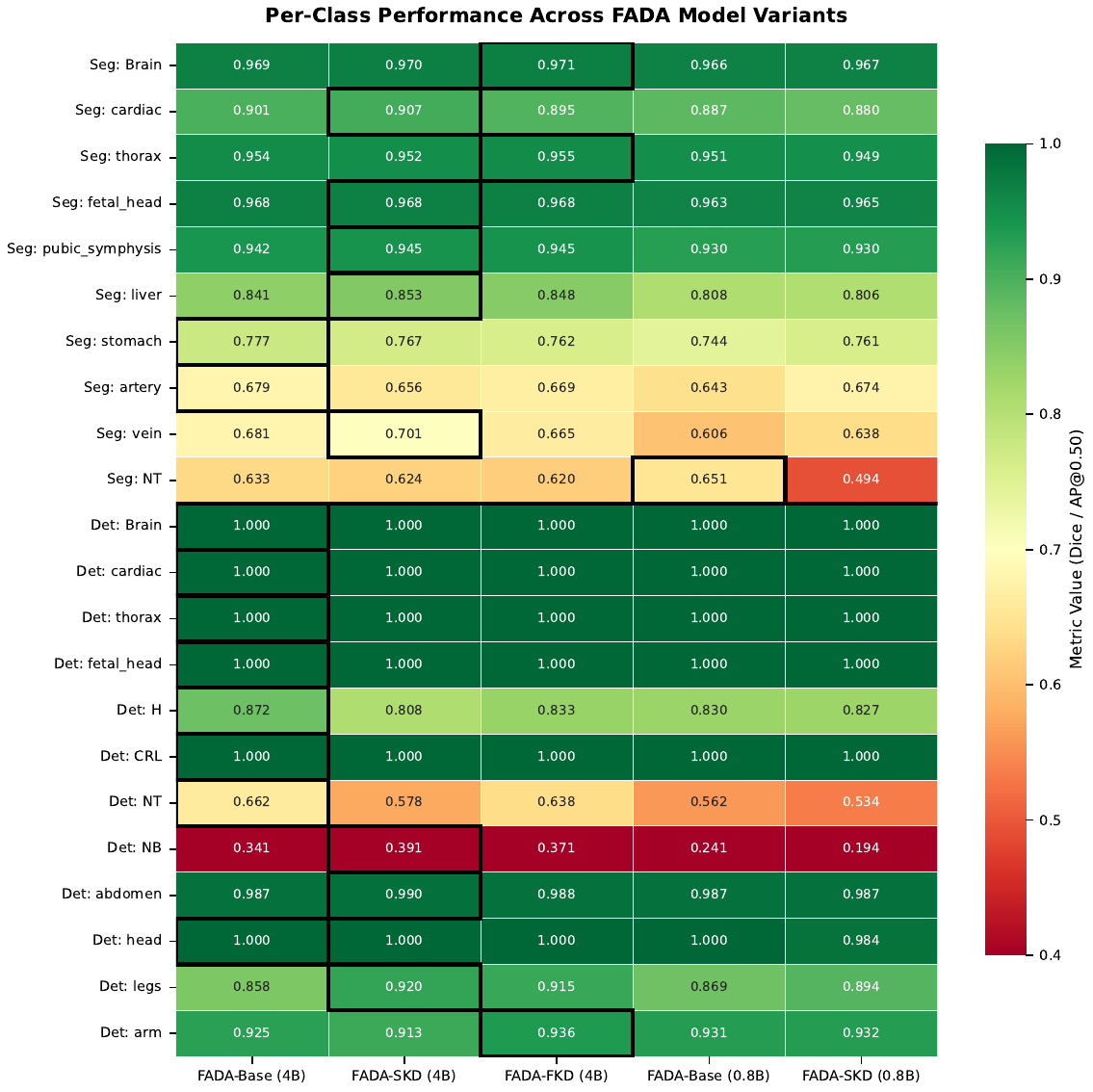}
    \caption{Per-class performance heatmap across FADA model variants. Left: detection AP@0.50 per anatomical class for each model variant. Right: segmentation Dice coefficient per class. Color intensity encodes metric value (darker = higher). All 4B models achieve near-perfect detection for large structures (Brain, Cardiac, Thorax). Performance differentiation emerges for fine structures: FADA-SKD leads in 7 of 10 segmentation classes while FADA-FKD excels at cavity septum detection.}
    \label{fig:heatmap}
\end{figure}

\paragraph{Per-class Analysis.}
All 4B variants achieve perfect or near-perfect detection for large anatomical structures (Brain, Cardiac, Thorax, Fetal Head: AP@0.50 $\geq$ 0.98), confirming robust detection across primary scan planes. Performance differences emerge for geometrically complex structures: FADA-FKD excels at cavity septum detection (CSP: 0.827 vs 0.795 SKD) while FADA-SKD achieves the highest segmentation Dice in 5 of 10 evaluated structures (cardiac, fetal head, liver, pubic symphysis, vein) and outperforms FADA-FKD in 7 of 10 classes, with notable gains on liver (+1.2\% vs Base) and vein (+1.9\% vs Base). Segmentation of thin membranous structures remains challenging across all variants (NT Dice: 0.620--0.633). Figure~\ref{fig:det_comparison} shows representative detection examples and Figure~\ref{fig:heatmap} provides a comprehensive per-class heatmap across all model variants.


\begin{figure}[!t]
    \centering
    \includegraphics[width=\textwidth]{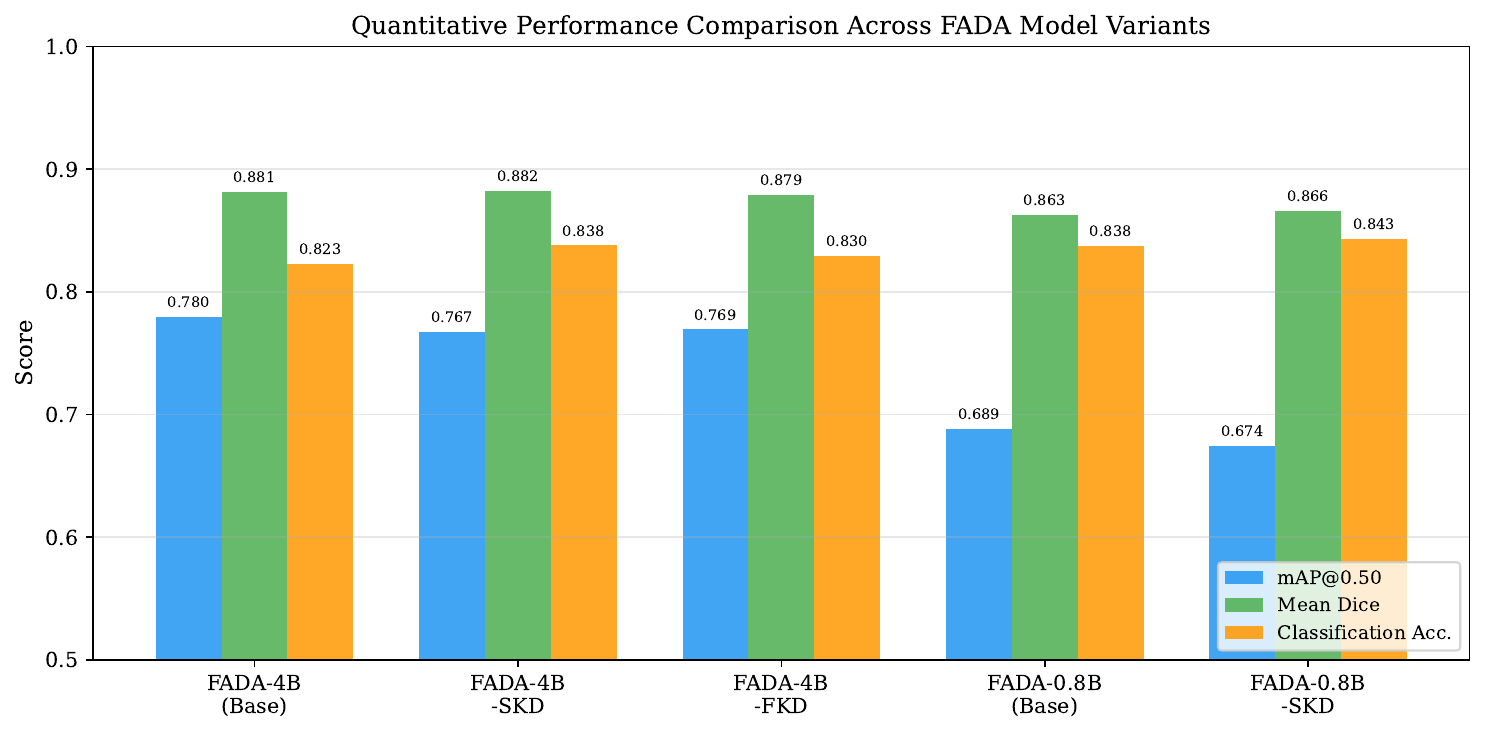}
    \caption{Performance comparison of FADA model variants across detection (mAP@0.50, mAP@0.75), segmentation (Dice, IoU), and classification accuracy metrics with 95\% bootstrap confidence intervals (error bars). FADA-SKD (4B) achieves the best overall balance: highest segmentation (Dice=0.882) and classification accuracy (0.838), while maintaining detection performance within the confidence interval of the Base model. The 0.8B variants retain 88--98\% of 4B performance across all metrics despite 5$\times$ fewer parameters.}
    \label{fig:model_comparison}
\end{figure}

\begin{figure}[!t]
    \centering
    \includegraphics[width=\textwidth]{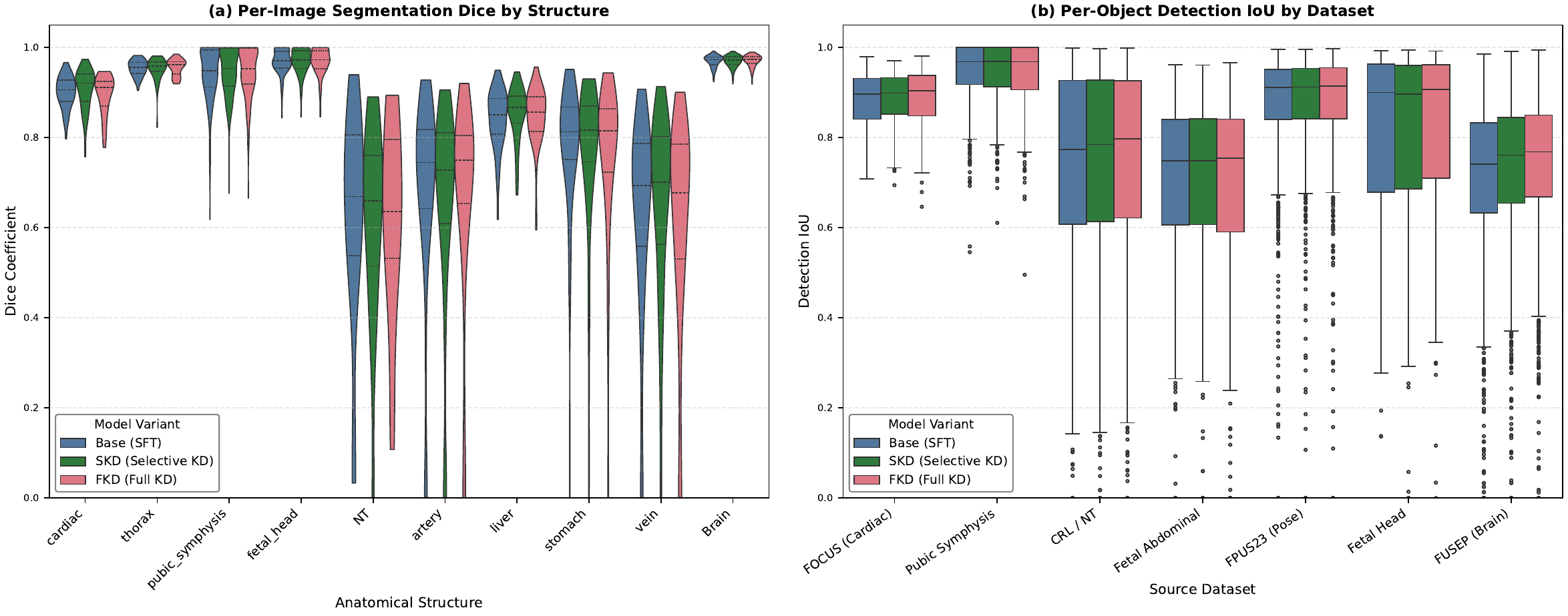}
    \caption{Score distribution analysis across model variants. Left: violin plots showing the distribution of per-sample segmentation Dice scores for each model variant (Base = SFT only, SKD = Selective KD, FKD = Full KD), with embedded box plots indicating median and interquartile range. Right: detection IoU score distributions by source dataset. FADA-SKD exhibits the tightest distribution with highest median Dice (0.882). The bimodal detection distributions reflect the dichotomy between easily detected large structures (IoU$>$0.8) and challenging fine structures.}
    \label{fig:distributions}
\end{figure}

\paragraph{Per-dataset Performance.}
Performance varies across source datasets, reflecting inherent task difficulty. The FOCUS cardiac dataset yields the highest aggregate scores (mAP@0.50: 1.00, Dice: 0.928) owing to well-defined structure boundaries in four-chamber views. Classification performance is notably higher on the Fetal Echocardiography dataset (0.904) than on FPUS23 (0.713), reflecting the former's more discriminative inter-class visual features versus the subtle postural differences in fetal pose classification. Figure~\ref{fig:model_comparison} presents an overview comparison across all models and tasks. The per-sample score distributions (Figure~\ref{fig:distributions}) further reveal that FADA-SKD exhibits the tightest distribution with the highest median, while supplementary Figures~S10--S12 provide additional keypoint detection, classification, and pairwise comparisons.

\begin{figure}[!t]
    \centering
    \includegraphics[width=\textwidth]{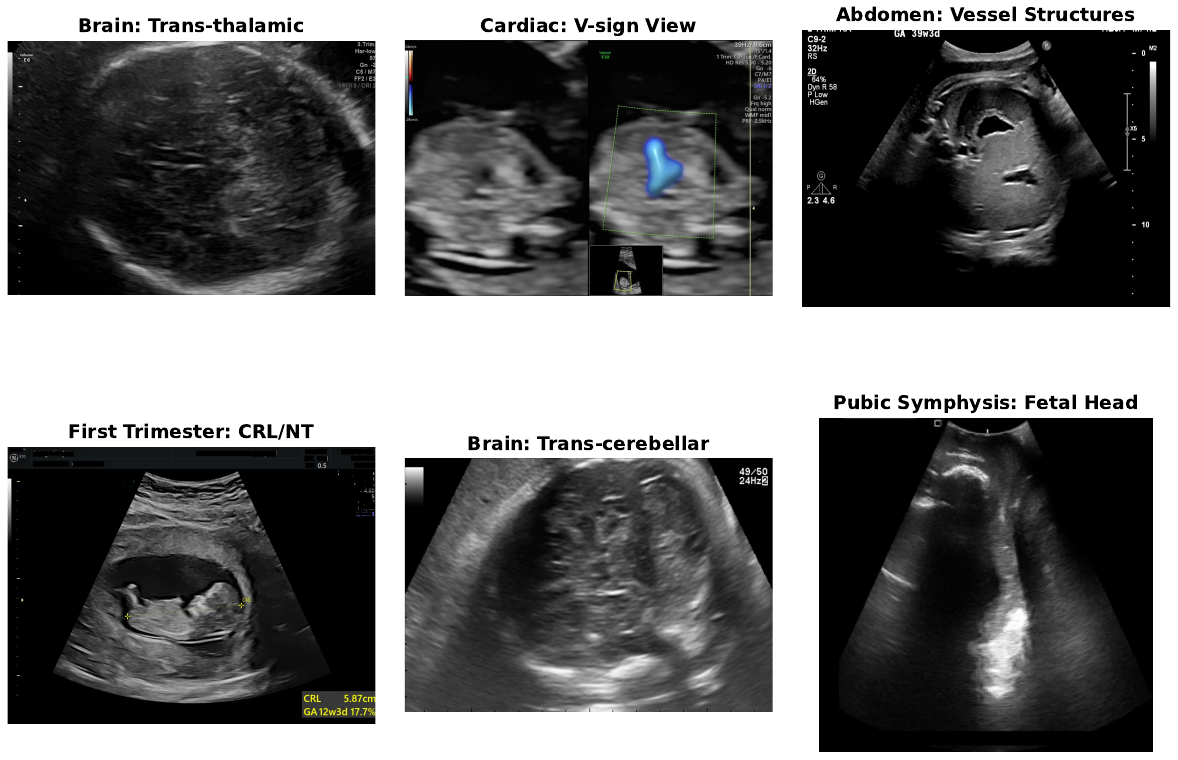}
    \caption{Representative fetal ultrasound images from the FADA evaluation dataset spanning diverse anatomical categories. The system processes each image through the interpretation-first pipeline, autonomously determining appropriate detection and segmentation targets based on identified anatomy. Top row (left to right): trans-thalamic brain view, cardiac V-sign view, abdominal vessel structures. Bottom row: first-trimester CRL/NT screening, trans-cerebellar brain view, pubic symphysis with fetal head.}
    \label{fig:qualitative}
\end{figure}

\subsection{Expert Sonographer Validation}\label{sec:expert_eval}

An expert sonographer independently evaluated all three 4B model variants on 237 images (62 external clinical images and 175 from the test set) spanning 18 anatomical categories using a blinded scoring protocol. For each image, the sonographer assigned a quality score from 1 (clinically acceptable, no correction needed) to 3 (poor, major errors) for both annotation quality (bounding box and segmentation mask accuracy) and interpretation quality (clinical correctness and completeness). Table~\ref{tab:sonographer} presents the results.

\begin{table}[!t]
\caption{Expert sonographer evaluation (n=237 images, blinded). Scores: 1\,=\,clinically acceptable, 2\,=\,partial errors, 3\,=\,failure (lower is better). Score distribution shown as percentage of images.}\label{tab:sonographer}
\begin{tabular*}{\textwidth}{@{\extracolsep\fill}lcccc}
\toprule
\textbf{Model} & \textbf{Score 1 (\%)} & \textbf{Score 2 (\%)} & \textbf{Score 3 (\%)} & \textbf{Mean} \\
\midrule
\multicolumn{5}{l}{\textit{Annotation quality (bounding boxes + segmentation masks)}} \\
FADA-Base & 36.3 & 25.7 & 38.0 & 2.017 \\
FADA-SKD  & 35.0 & 27.4 & 37.6 & 2.025 \\
FADA-FKD  & 33.3 & 28.3 & 38.4 & 2.051 \\
\midrule
\multicolumn{5}{l}{\textit{Interpretation quality (structured clinical text)}} \\
FADA-Base & 29.5 & 30.0 & 40.5 & 2.110 \\
FADA-SKD  & \textbf{38.0} & 31.6 & \textbf{30.4} & \textbf{1.924} \\
FADA-FKD  & 27.4 & 27.0 & 45.6 & 2.181 \\
\bottomrule
\end{tabular*}
\end{table}

FADA-SKD achieves the best interpretation score (mean 1.924) and the highest proportion of clinically acceptable outputs (38.0\% Score\,=\,1) compared to FADA-Base (29.5\%) and FADA-FKD (27.4\%). SKD also records the lowest failure rate for interpretation (30.4\% vs 40.5\% for Base), supporting the hypothesis that selective distillation preserves the language model's clinical reasoning while benefiting annotation quality through indirect knowledge transfer. Annotation scores remain comparable across all three variants (mean 2.017--2.051), consistent with quantitative metrics showing similar detection performance.

The discrepancy between FADA-FKD's higher automated classification accuracy (Table~\ref{tab:main_results}) and its lower human interpretation score suggests that feature alignment on interpretation data may improve pattern matching for classification labels while degrading free-text clinical reasoning quality.

\subsection{Human-in-the-Loop Evaluation}\label{sec:hil}

To assess FADA-SKD under realistic clinical deployment conditions, a human-in-the-loop (HiL) evaluation was conducted using the deployed web application. An expert sonographer processed 49 clinical cases, with the ability to select specific analysis phases and provide corrective feedback. Table~\ref{tab:hil} presents the scoring results.

\begin{table}[!t]
\caption{Human-in-the-Loop evaluation of FADA-SKD deployed in the web application. An expert sonographer scored 49 clinical cases on the same 1--3 scale. Score distribution (percentage of cases) is shown.}\label{tab:hil}
\begin{tabular*}{\textwidth}{@{\extracolsep\fill}lccccc}
\toprule
\textbf{Task} & \textbf{Mean Score} & \textbf{Score 1 (\%)} & \textbf{Score 2 (\%)} & \textbf{Score 3 (\%)} \\
\midrule
Interpretation & \textbf{1.286} & 73.5 & 24.5 & 2.0 \\
Annotation     & 1.449 & 63.3 & 28.6 & 8.2 \\
\bottomrule
\end{tabular*}
\end{table}

In HiL mode, FADA-SKD achieves substantially better scores than in fully autonomous evaluation (interpretation: 1.286 vs 1.924; annotation: 1.449 vs 2.025), with 73.5\% of interpretations receiving a perfect score and only 2.0\% rated as poor. This improvement reflects the interactive deployment where sonographers guide the analysis pipeline by selecting specific phases, view types, and detection targets, thereby reducing error propagation from the interpretation-first cascade. The data confirm that FADA can function as an effective clinical decision support tool when paired with even minimal operator expertise.

\subsection{Training Dynamics}\label{sec:training_dynamics}

\begin{figure}[!t]
    \centering
    \includegraphics[width=\textwidth]{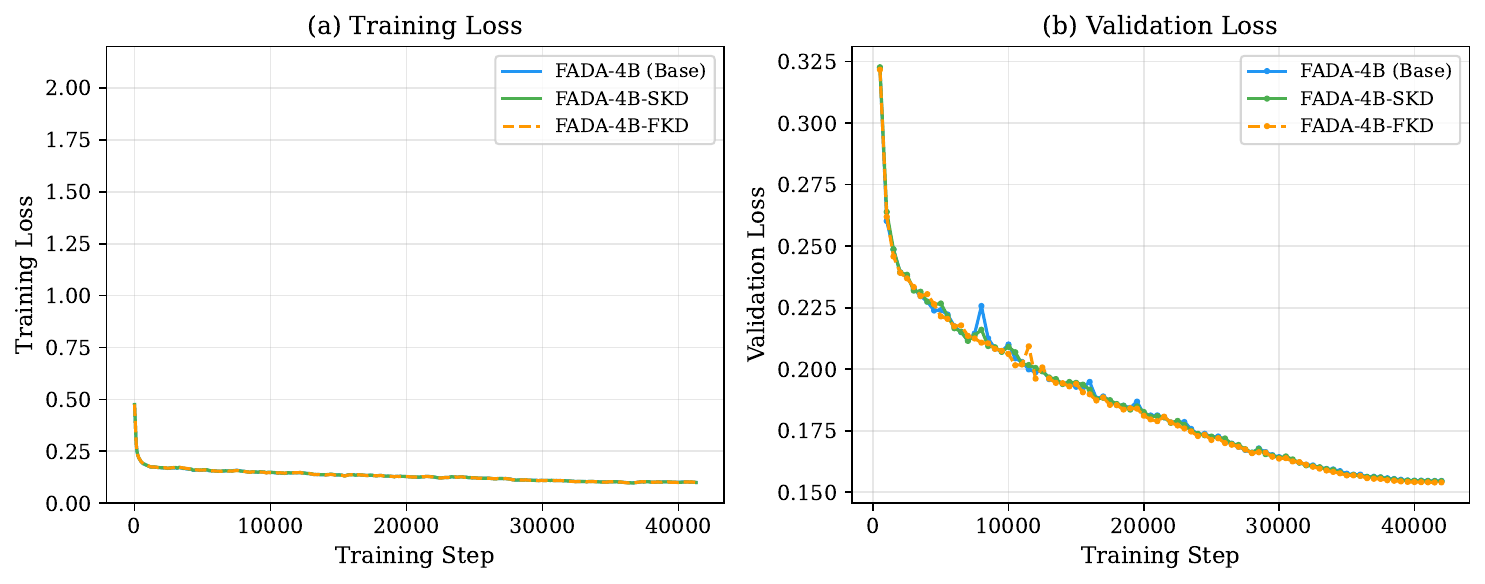}
    \caption{Training dynamics for FADA 4B model variants over 3 epochs (42,285 steps). (a)~Smoothed training loss (window=100 steps): all variants converge to similar final loss ($\approx$0.12), with SKD and FKD showing marginally faster initial convergence due to additional gradient signal from feature alignment. (b)~Validation loss evaluated every 500 steps: final validation loss $\approx$0.155 across all variants, confirming that the distillation objective does not impede generalization or introduce overfitting.}
    \label{fig:training_loss}
\end{figure}

Figure~\ref{fig:training_loss} shows training and validation loss curves for all three 4B model variants. All models converge to similar final training loss ($\approx$0.12) and validation loss ($\approx$0.155) after 3 epochs (42,285 steps), indicating that the distillation objective does not impede convergence. The SKD and FKD variants exhibit marginally faster initial convergence compared to the base model, consistent with feature alignment providing additional gradient signal during early training.

\subsection{Interpretability Analysis}\label{sec:xai}

Attention-based interpretability is increasingly recognized as essential for clinical trust in medical VLMs. FetalMind~\cite{he2025fetalmind} showed that attention to disease-relevant views correlates positively with diagnostic accuracy, while Ultrasound-CLIP~\cite{jin2026ultrasoundclip} demonstrated that structured diagnostic attributes improve clinical reasoning. We adopt complementary interpretability analyses here to explain \emph{why} FADA-SKD produces superior clinical outputs despite receiving no feature-level supervision on interpretation data.

\begin{figure}[H]
    \centering
    \includegraphics[width=\textwidth]{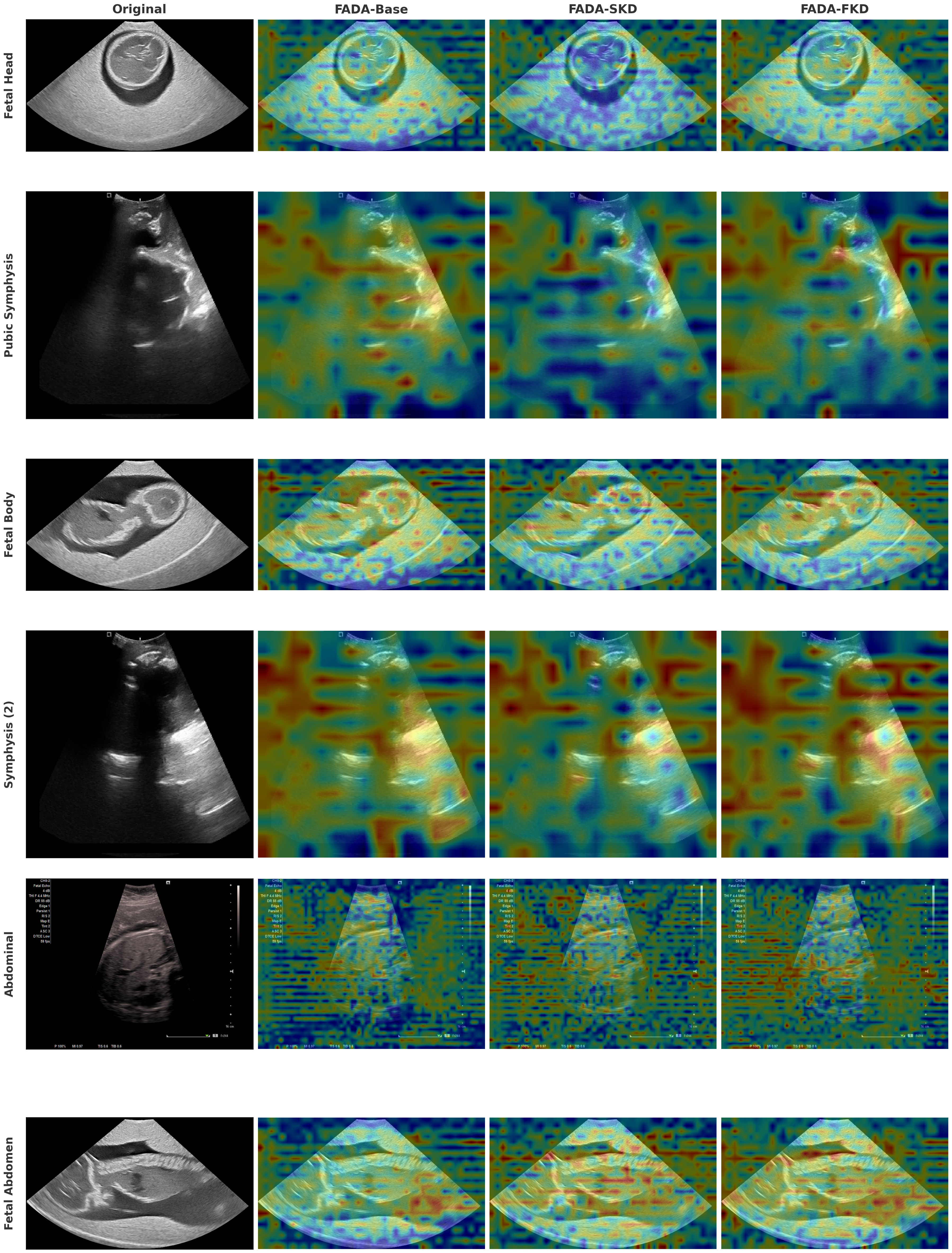}
    \caption{Vision encoder attention heatmaps (layer 23) across FADA variants for 6 test-set images selected via quantitative focus analysis. Rows: fetal head, pubic symphysis (2 views), fetal body, fetal abdominal structures, and fetal abdomen. Columns show the original ultrasound image followed by attention overlays from FADA-Base, FADA-SKD, and FADA-FKD. Color scale: blue (low) to yellow/red (high attention). FADA-SKD concentrates attention on diagnostically relevant anatomical structures, including the fetal skull ring (row~1), symphysis landmarks (rows~2,~4), and fetal body boundaries (row~3), while FADA-FKD exhibits more diffuse spatial patterns with scattered hot spots extending toward image periphery. This pattern is consistent with teacher-forced structural alignment pulling attention toward spatial boundaries rather than semantically informative regions.}
    \label{fig:attention}
\end{figure}

\paragraph{Attention Pattern Analysis.}
Attention heatmaps from the vision encoder's final layer across model variants (Figure~\ref{fig:attention}) reveal that FADA-SKD consistently focuses on clinically relevant anatomical landmarks: cardiac chambers in four-chamber views, femur boundaries in biometry planes, and ventricular structures in brain views. FADA-FKD, by contrast, displays more diffuse attention patterns extending to image periphery. This observation is consistent with our hypothesis: full distillation forces spatial alignment during interpretation, pulling attention toward structural boundaries (optimized for detection) rather than diagnostically informative regions (needed for clinical reasoning). Selective KD avoids this conflict, allowing the model to develop attention patterns naturally suited to each task type.

\paragraph{Structured Output Quality.}
Token-level attribution analysis of interpretation outputs (Supplementary Table~S7) reveals that FADA-SKD achieves the highest per-field semantic accuracy (mean 0.753 across 8 JSON fields vs 0.738 for Base and 0.744 for FKD), the highest clinical terminology density (17.3 clinical terms per output vs 17.0 for Base and 17.1 for FKD), and the most anatomical structures correctly identified per image (3.96 vs 3.72 Base). All three variants achieve 100\% JSON field completeness. FADA-FKD notably degrades on BLEU-1 (0.752 vs 0.766 for Base/SKD) and ROUGE-L (0.774 vs 0.790), confirming that full distillation introduces noise into the language generation pathway. These findings parallel Ultrasound-CLIP's emphasis on structured diagnostic attributes: models that preserve clinical attribute generation integrity produce more trustworthy outputs.

\begin{figure}[H]
    \centering
    \includegraphics[width=\textwidth]{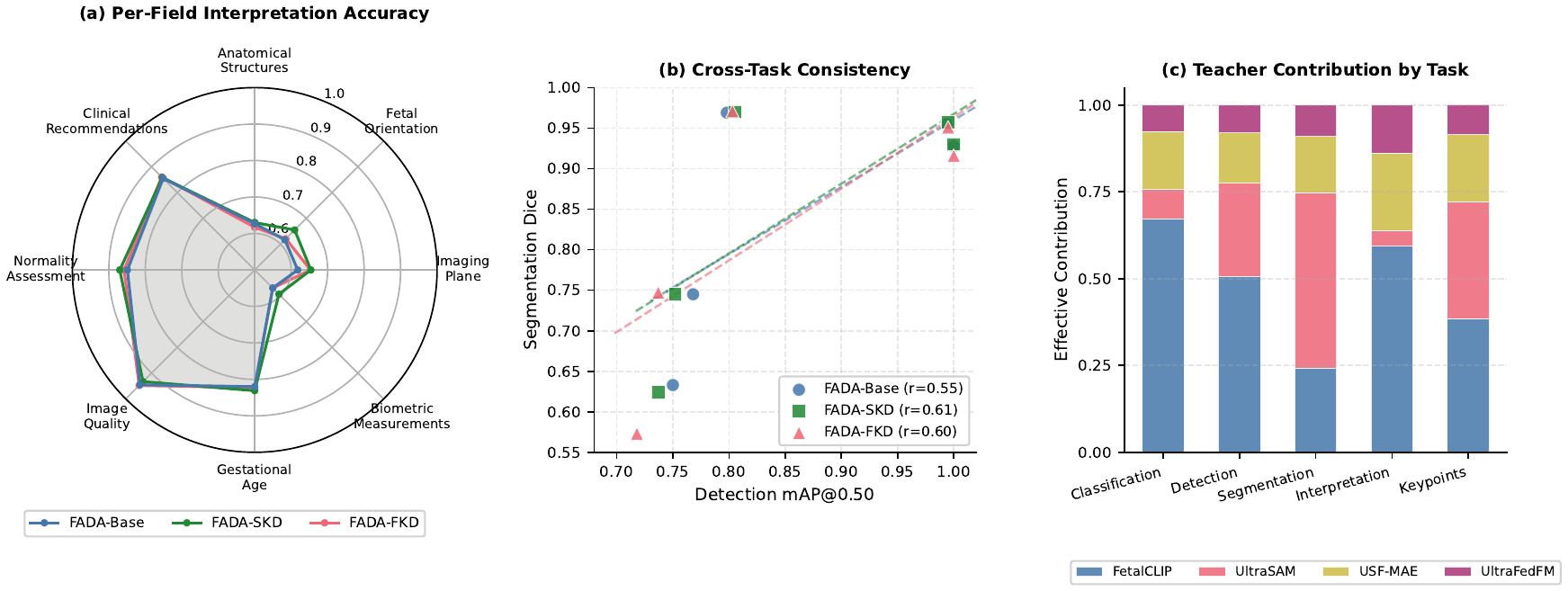}
    \caption{Interpretability analysis of FADA model variants. (a)~Per-field semantic accuracy across the 8 JSON interpretation fields: FADA-SKD (green) leads in 5 of 8 fields, with particular advantages in clinically critical fields (fetal orientation, imaging plane, biometric measurements). (b)~Cross-task consistency: detection mAP@0.50 vs segmentation Dice per dataset, showing FADA-FKD achieves the tightest detection--segmentation coupling (Pearson $r$=0.74 vs 0.61 SKD, 0.55 Base). (c)~Effective teacher contribution by task type: FetalCLIP dominates interpretation (59\%) and classification (67\%), while UltraSAM dominates segmentation (50\%), explaining why selective KD (which applies teacher features only to annotation tasks) preserves interpretation quality.}
    \label{fig:xai_summary}
\end{figure}

\paragraph{Cross-Task Consistency.}
The detection-segmentation Pearson correlation (Section~\ref{sec:quantitative}) provides an additional interpretability signal: FADA-FKD achieves $r$=0.74, followed by FADA-SKD at $r$=0.61 and Base at $r$=0.55 (Figure~\ref{fig:xai_summary}b). While FKD exhibits the tightest detection--segmentation coupling, FADA-SKD remains the recommended deployment variant because its slightly lower spatial correlation accompanies substantially better interpretation quality and expert ratings, properties more critical for clinical utility than internal metric coherence alone.

\paragraph{Interpretability Summary.}
Taken together, these analyses support FADA-SKD as the recommended deployment variant. It achieves (1) focused clinical attention patterns, (2) the highest structured interpretation quality, (3) superior expert ratings, and (4) the strongest internal consistency across spatial tasks. The selective KD strategy succeeds because it applies spatial teacher expertise only where beneficial (annotation tasks), preserving the language model's capacity for nuanced clinical reasoning on interpretation tasks.


\section{Discussion}\label{sec:discussion}

\paragraph{Why Selective KD Outperforms Full KD.}
The central finding of this study is that selective knowledge distillation, applying feature alignment exclusively to annotation data, consistently outperforms full distillation across segmentation, expert-rated interpretation quality, and strict detection thresholds. This result can be attributed to the nature of the four teacher models (FetalCLIP, UltraSAM, USF-MAE, UltraFedFM), which encode visual-spatial patterns optimized for structural recognition and localization. When these features are aligned with the student during interpretation training, they introduce conflicting gradients: the feature loss encourages spatial feature patterns while the language modeling loss requires abstract clinical reasoning over the full image context. Selective KD resolves this conflict by allowing interpretation training to optimize language generation without spatial feature constraints, while annotation training benefits from the teachers' spatial expertise.

The same pattern appears in the multi-task learning literature, where auxiliary objectives improve performance only when they share relevant inductive biases with the primary task~\cite{gou2021knowledge}. It also complements the structured evaluation decomposition in FetalMind~\cite{he2025fetalmind}, which similarly recognizes that different aspects of clinical assessment benefit from distinct supervisory signals. Teacher contribution analysis (Supplementary Table~S6, Figure~\ref{fig:xai_summary}c) reveals that FetalCLIP dominates interpretation feature alignment (59\% effective weight) due to its contrastive vision-language pre-training, while UltraSAM dominates segmentation (50\%) through spatial SAM-based features. This task-specific teacher dominance explains why applying all teacher features uniformly (FKD) degrades interpretation: UltraSAM's spatial features conflict with the linguistic reasoning required for clinical text generation. A comprehensive ablation study (Supplementary Materials) confirms that multi-teacher fusion provides +2.2\% mAP over single-teacher distillation, that cosine similarity loss collapses feature diversity ($-$3.1\% mAP), and that VL pre-training contributes more than KD alone ($-$4.5\% mAP when training from scratch).

\paragraph{Cross-task Knowledge Transfer.}
FADA-SKD generates clinically meaningful interpretations for datasets appearing only in annotation training (e.g., FUSEP brain anatomy, FOCUS cardiac structures). Despite never encountering interpretation examples for these specific datasets, the model produces accurate 8-field JSON assessments including correct anatomical structure identification, imaging plane determination, and normality assessment. This suggests effective knowledge transfer from annotation supervision to interpretation capability, likely mediated by the shared visual encoder that learns generalizable anatomical features during detection and segmentation training.

\paragraph{Comparison with Related Work.}
Unlike Ultrasound-CLIP~\cite{jin2026ultrasoundclip}, which employs heterogeneous graph encoding and semantic soft labels for ultrasound-text alignment, FADA uses a generative VLM approach that directly produces structured clinical outputs rather than embedding-space matching. This enables more expressive and clinically actionable outputs at the cost of requiring more training data. Compared to FetalMind~\cite{he2025fetalmind}, which uses disease-view bipartite graphs and separate evaluation diagnostics for different assessment aspects, FADA handles all analysis tasks within a single autoregressive generation process, simplifying deployment while maintaining competitive performance.

Relative to SonoMate~\cite{guo2025sonomate}, which achieves strong performance through video-level grounding in Nature Biomedical Engineering, FADA operates on single images, a critical distinction for point-of-care settings where real-time video capture and storage infrastructure may be unavailable. FADA also addresses five concurrent tasks (interpretation, classification, detection, segmentation, keypoints) whereas SonoMate focuses on detection and report generation. Compared to MobileFetalCLIP~\cite{saeed2025mobilefetalclip}, which compresses FetalCLIP for mobile classification using DARK (Diagonal-Anchored Repulsive KD), FADA's selective KD operates across four heterogeneous teachers and preserves multi-task generative capability rather than reducing to classification embeddings. The approach also benefits from task-conditional distillation, a strategy not explored in existing medical KD frameworks such as ClinKD~\cite{chen2025clinkd} or MoVE-KD~\cite{cao2025movekd}, which apply uniform distillation across all training samples.

General-purpose medical VLMs (e.g., GPT-4V, LLaVA-Med) have been benchmarked on fetal ultrasound interpretation with generally poor performance on domain-specific tasks such as biometric measurement identification and anatomical orientation assessment. FADA's domain-specific training on 56,805 interpretation conversations and 12,000 annotated images enables structured clinical outputs that general-purpose models cannot reliably produce, particularly the normalized coordinate outputs required for detection and segmentation overlays. Unlike API-based models, FADA's on-premise deployment avoids patient data privacy concerns inherent in cloud-based inference for medical imaging.

\paragraph{Clinical Implications for LMICs.}
The human-in-the-loop evaluation demonstrates that FADA functions effectively as a clinical decision support tool, with 73.5\% of interpretations requiring no correction. This mode is particularly relevant for LMICs, where community health workers or general practitioners may perform ultrasound screening without specialized sonography training. The system's capacity to autonomously determine appropriate detection and segmentation targets without requiring class label input removes a key barrier to deployment where diagnostic expertise is unavailable.

\begin{figure}[H]
    \centering
    \begin{subfigure}{0.24\textwidth}
        \centering
        \includegraphics[width=\textwidth]{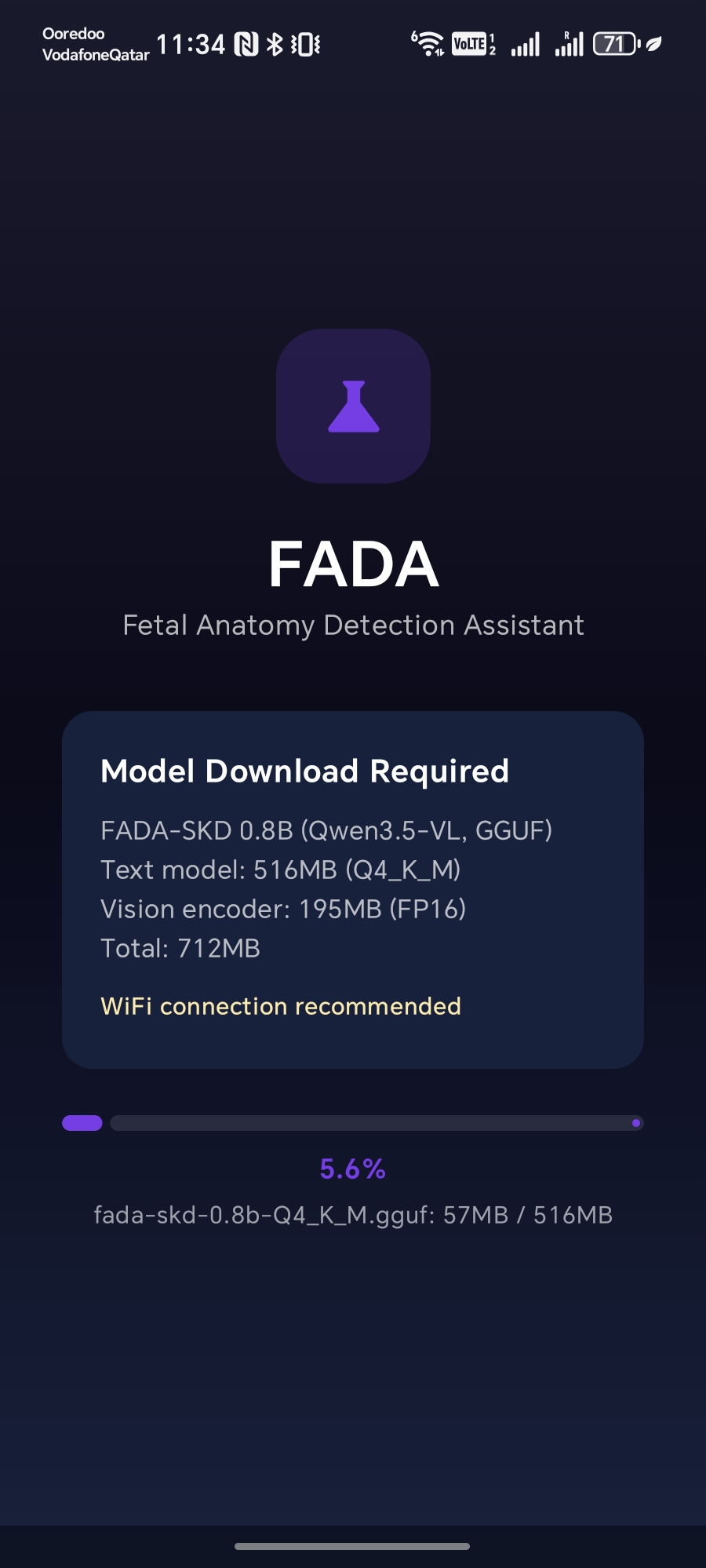}
        \caption{Model download}
    \end{subfigure}
    \hfill
    \begin{subfigure}{0.24\textwidth}
        \centering
        \includegraphics[width=\textwidth]{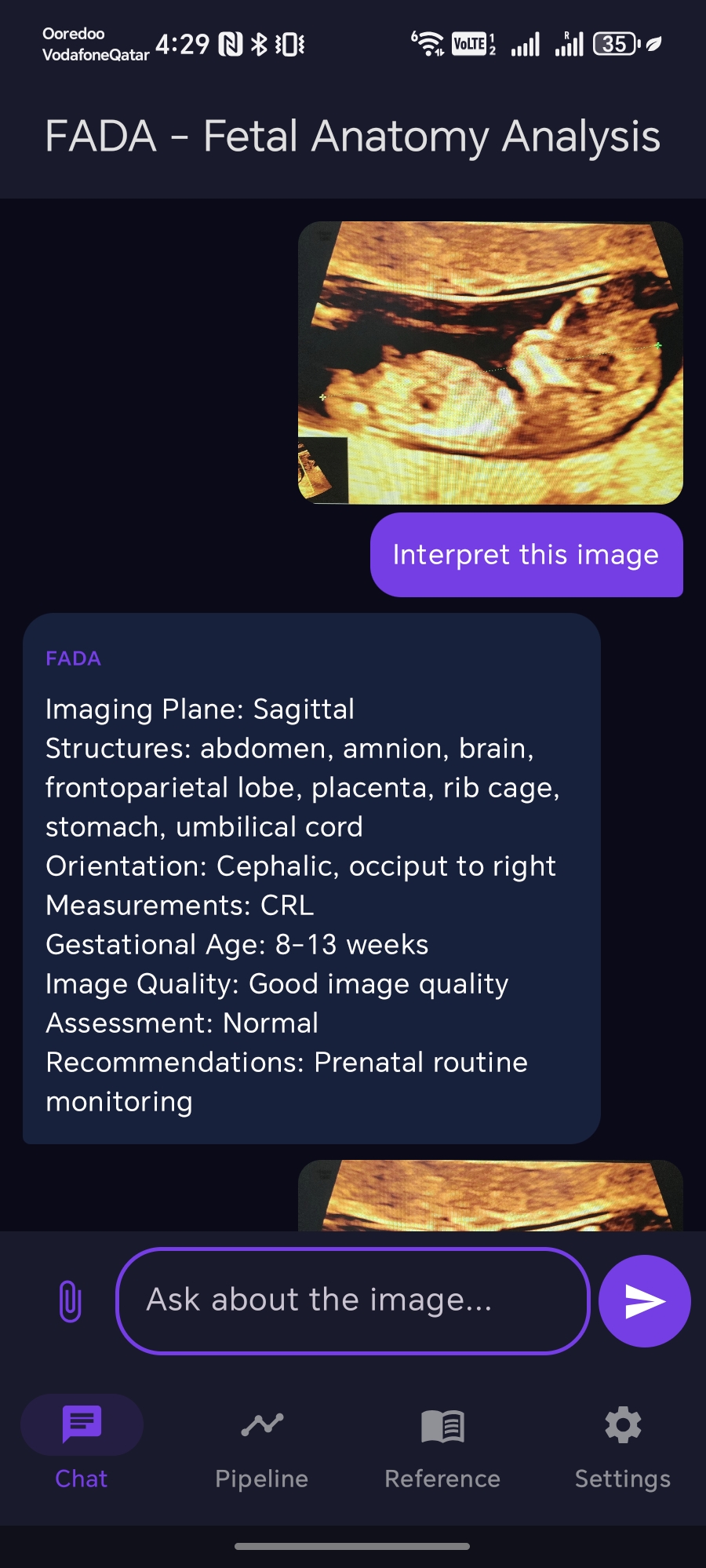}
        \caption{Interpretation}
    \end{subfigure}
    \hfill
    \begin{subfigure}{0.24\textwidth}
        \centering
        \includegraphics[width=\textwidth]{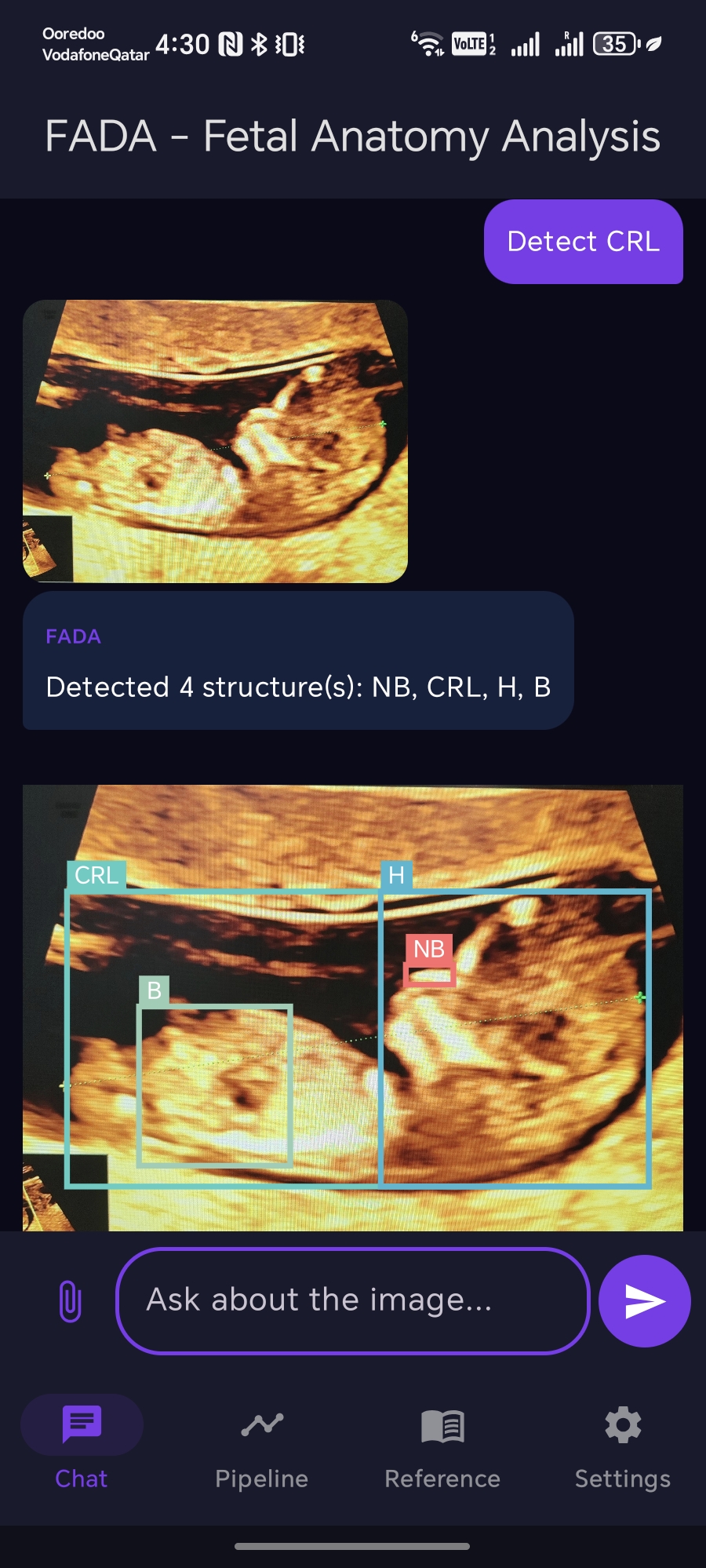}
        \caption{Detection overlay}
    \end{subfigure}
    \hfill
    \begin{subfigure}{0.24\textwidth}
        \centering
        \includegraphics[width=\textwidth]{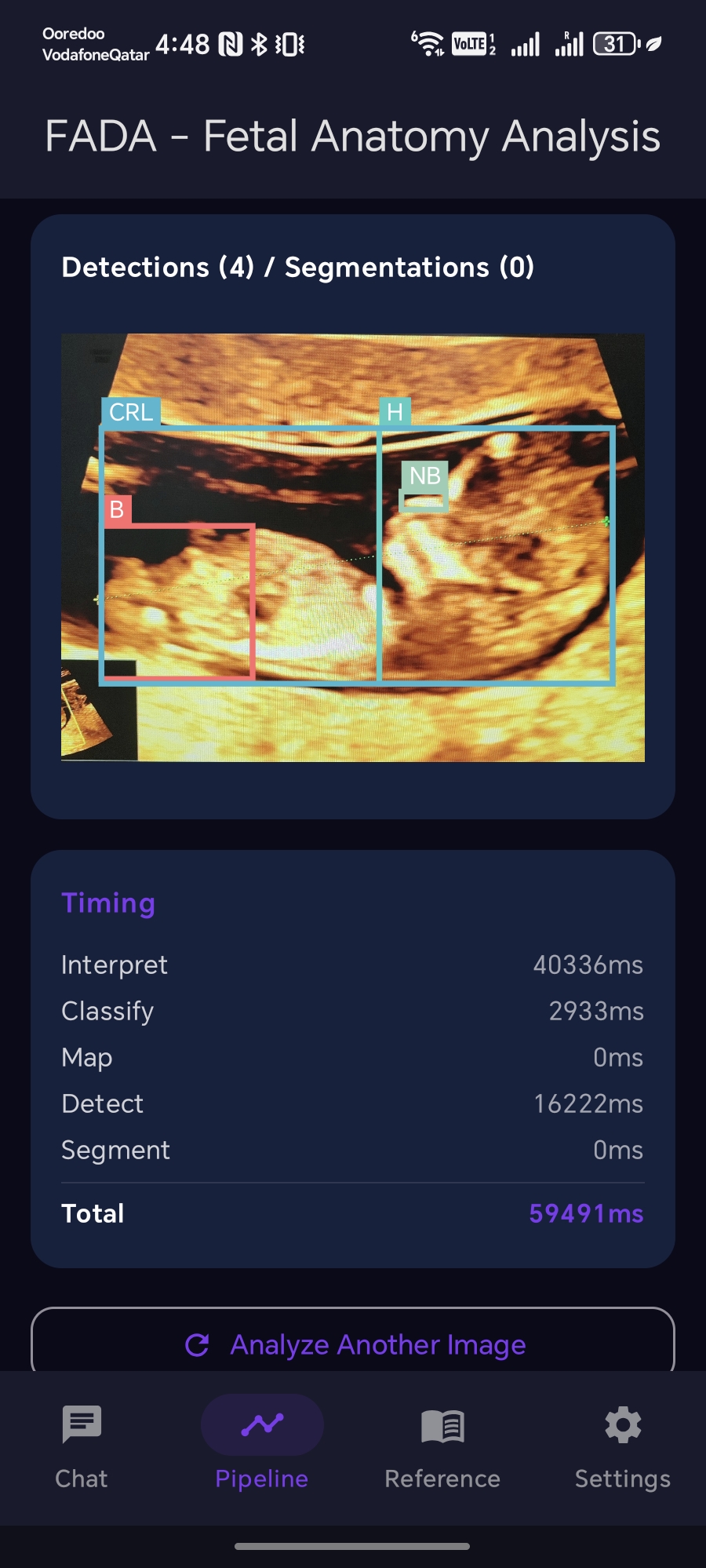}
        \caption{Auto pipeline}
    \end{subfigure}
    \caption{FADA mobile application deployed on a commodity Android smartphone (Honor~90, Snapdragon~7 Gen~1, 12\,GB RAM) running entirely offline via llama.cpp with GGUF quantization (Q4\_K\_M). (a)~One-time model download (712\,MB total: 516\,MB text model + 195\,MB vision encoder). (b)~Chat-mode interpretation: the user attaches a fetal ultrasound image and receives a structured clinical assessment (${\sim}$40\,s per task). (c)~Detection with bounding-box overlay rendered on-device, identifying CRL, head (H), body (B), and nasal bone (NB). (d)~Autonomous 5-phase pipeline with per-phase timing (total ${\sim}$59\,s), demonstrating full offline operation without cloud connectivity.}
    \label{fig:mobileapp}
\end{figure}

The 0.8B model variants further demonstrate that knowledge distillation enables effective compression to edge-deployable scales (segmentation Dice: 0.866, classification accuracy: 0.843) while maintaining clinically useful performance. This opens pathways for offline deployment on portable ultrasound devices in remote facilities without reliable internet connectivity. Table~\ref{tab:deployment} summarizes deployment configurations and inference characteristics. To validate this pathway concretely, the 0.8B FADA-SKD model is quantized to GGUF format (Q4\_K\_M, 516\,MB text model + 195\,MB FP16 vision encoder; 712\,MB total) and deployed via llama.cpp on a commodity Android smartphone (Honor~90, Qualcomm Snapdragon~7 Gen~1, 12\,GB RAM, Android~15) without any cloud connectivity.\footnote{\url{https://huggingface.co/mshz88/FADA-Mobile-GGUF}} Figure~\ref{fig:mobileapp} shows the application interface. The full 5-phase autonomous pipeline completes in approximately 59\,s, with individual chat-mode tasks (interpretation or detection) completing in ${\sim}$40\,s. This demonstrates that the distilled model can be integrated with portable fetal ultrasound devices in a stand-alone fashion, enabling AI-assisted screening in facilities without internet infrastructure.

\begin{table}[!t]
\caption{FADA deployment configurations and inference characteristics. Latency is reported for the full 5-phase pipeline (interpretation + classification + detection + segmentation + keypoints) per image.$^\dagger$}\label{tab:deployment}
\begin{tabular*}{\textwidth}{@{\extracolsep\fill}lcccc}
\toprule
\textbf{Configuration} & \textbf{Model} & \textbf{Hardware} & \textbf{Latency}$^\dagger$ & \textbf{Connectivity} \\
\midrule
Cloud (high accuracy) & 4B FP16 & RTX 4090 (24 GB) & $\sim$20--25 s & Required \\
Cloud (cost-efficient) & 4B FP16 & A10G (24 GB) & $\sim$35--45 s & Required \\
Cloud (budget) & 4B FP16 & T4 (16 GB) & $\sim$70--140 s & Required \\
Edge (offline) & 0.8B Q4\_K\_M & Android (12 GB RAM) & $\sim$59 s & None \\
\bottomrule
\end{tabular*}
\vspace{2pt}
{\footnotesize $^\dagger$GPU latency estimated from evaluation pipeline timing (14,004\,s for 4,478 images on RTX~4090) scaled by hardware throughput ratios. Mobile latency measured on Honor~90 (Snapdragon~7 Gen~1, 12\,GB RAM) with llama.cpp GGUF inference.}
\end{table}

\paragraph{Clinical Implementation in Resource-Constrained Settings.}
The design of FADA reflects the practical realities of healthcare delivery in LMICs, addressing multiple deployment constraints simultaneously. The entire training pipeline operates on a single consumer GPU (NVIDIA RTX~4090, 24~GB VRAM) with approximately 40 hours per model variant, making reproduction and local adaptation feasible for research institutions without high-performance computing clusters. This contrasts with most foundation model approaches that require multi-GPU training infrastructure costing orders of magnitude more~\cite{recent2025advances}.

\begin{figure}[H]
    \centering
    \begin{subfigure}{0.32\textwidth}
        \centering
        \includegraphics[width=\textwidth]{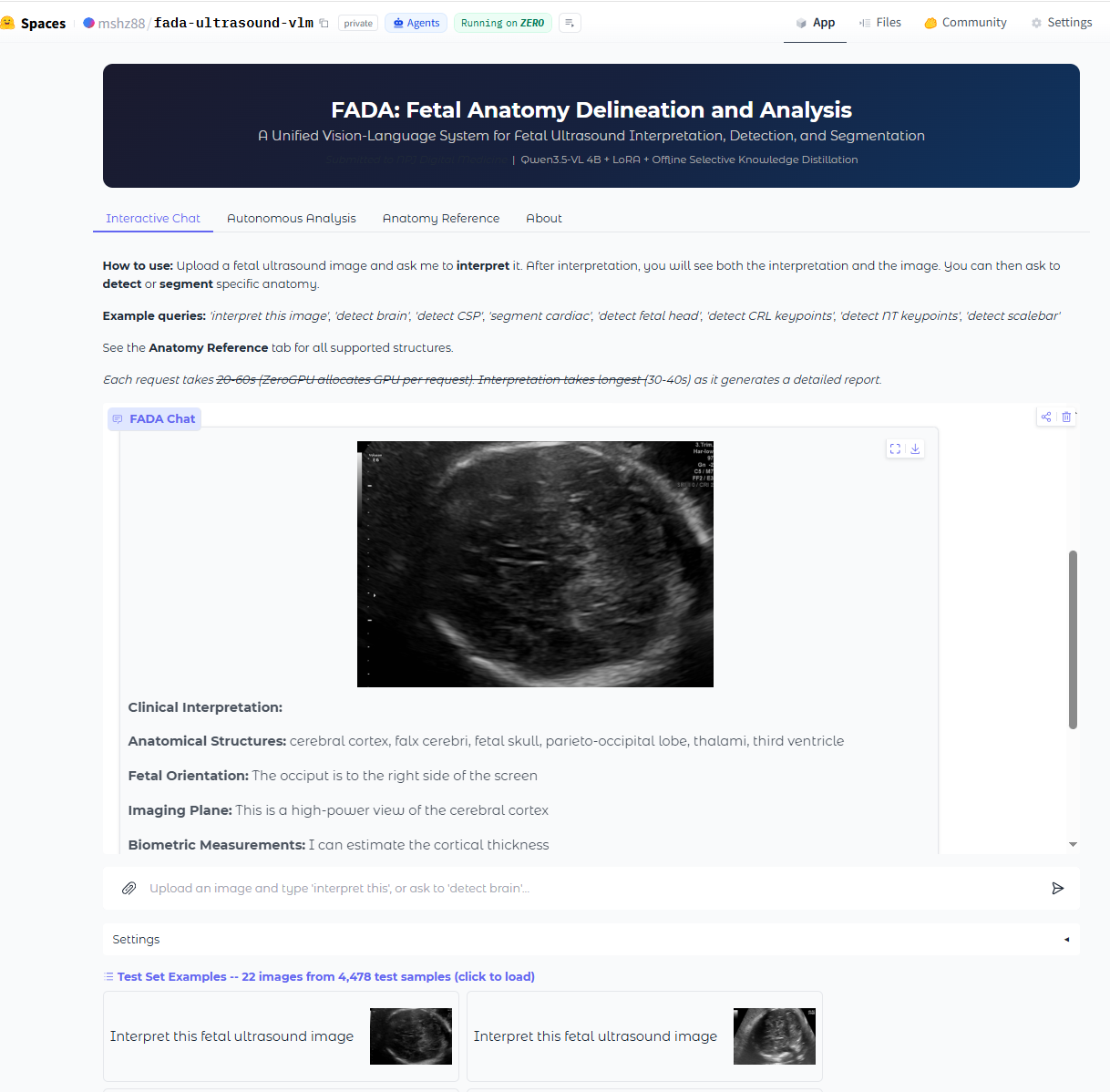}
        \caption{Human-in-the-loop}
    \end{subfigure}
    \hfill
    \begin{subfigure}{0.32\textwidth}
        \centering
        \includegraphics[width=\textwidth]{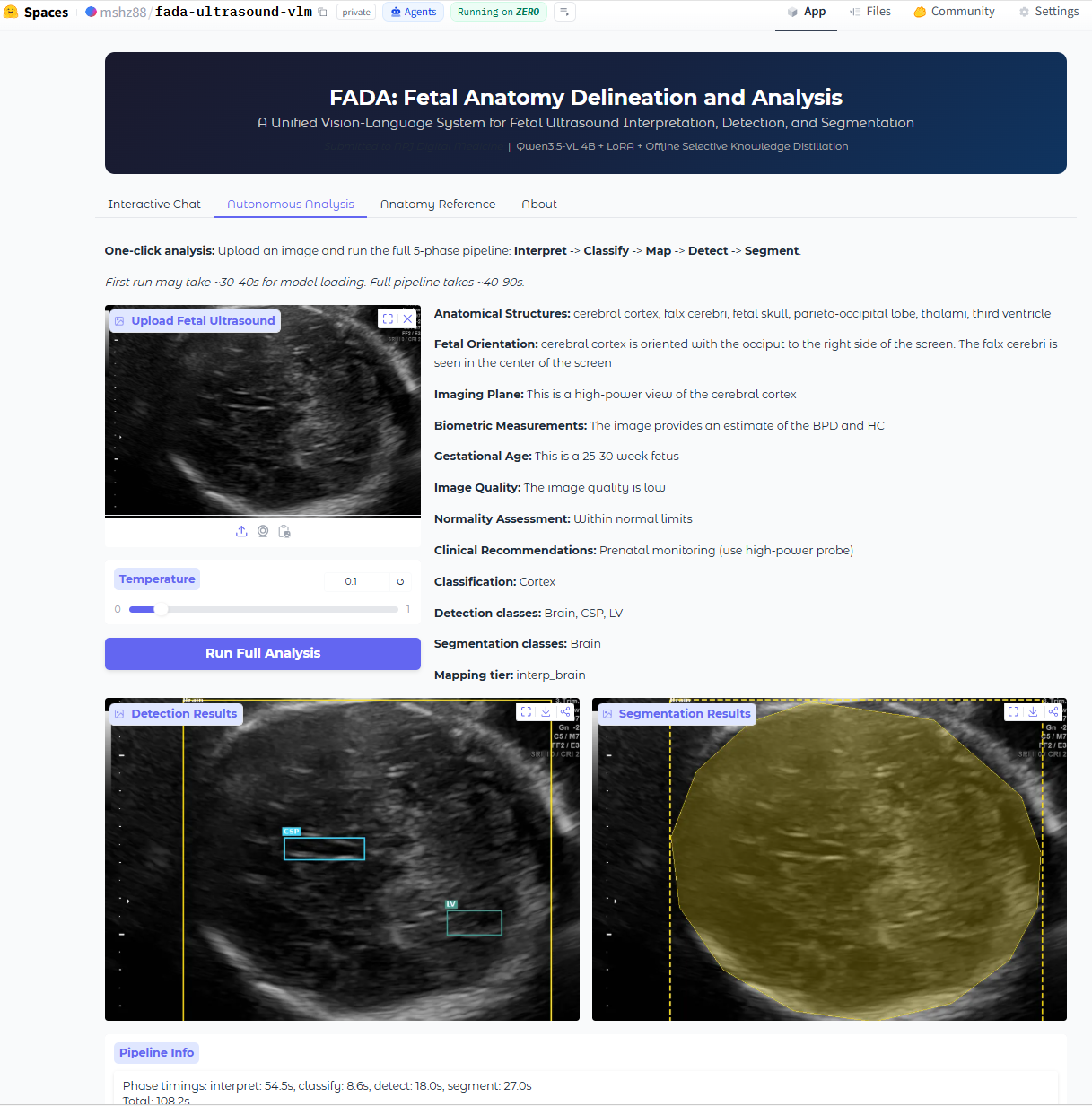}
        \caption{Autonomous mode}
    \end{subfigure}
    \hfill
    \begin{subfigure}{0.32\textwidth}
        \centering
        \includegraphics[width=\textwidth]{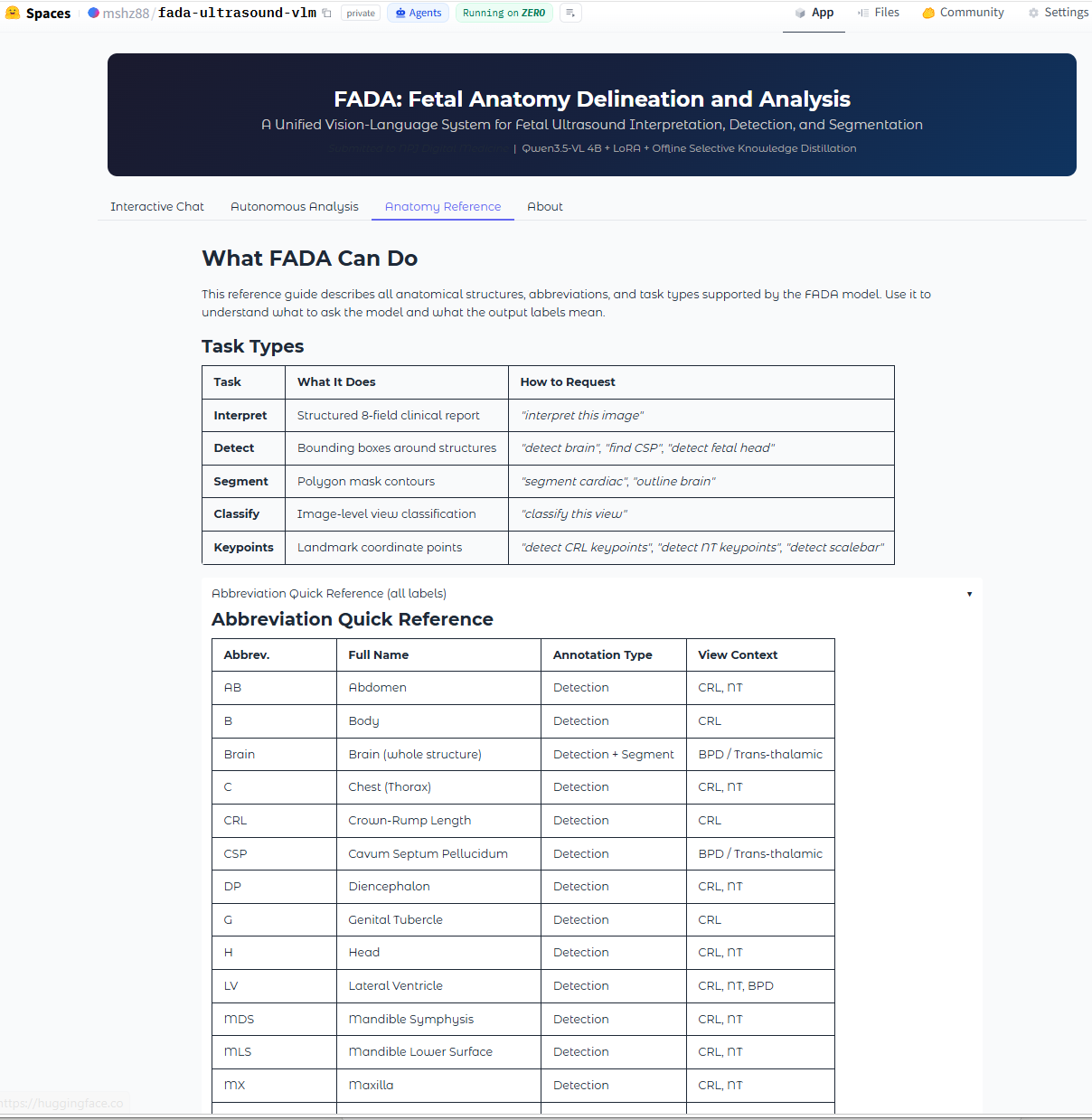}
        \caption{Reference view}
    \end{subfigure}
    \caption{FADA web application deployment interface. (a)~Human-in-the-loop mode: an expert sonographer reviews the initial interpretation and selectively guides subsequent analysis phases through an interactive chat interface. (b)~Autonomous mode: the system processes the uploaded ultrasound image through the full 5-phase pipeline without user intervention, producing structured interpretation, detection overlays, and segmentation masks. (c)~Reference documentation view: clinical reference information supporting operator decision-making in resource-constrained settings.}
    \label{fig:webapp}
\end{figure}

The open-source web application (Figure~\ref{fig:webapp}) requires no specialized hardware for inference: a standard server with a single mid-range GPU can serve multiple concurrent users, while the 0.8B edge model enables fully offline operation on portable devices (Figure~\ref{fig:mobileapp}). This dual-mode architecture supports tiered deployment where connected facilities use the 4B cloud model for maximum accuracy and remote clinics without reliable connectivity use the 0.8B model for autonomous screening with periodic synchronization for expert review.

The human-in-the-loop design enables a task-shifting workflow where non-specialist operators (community health workers, nurses, general practitioners) perform ultrasound acquisition while the AI provides immediate structured assessment. Remote expert sonographers can then review flagged cases asynchronously, effectively multiplying specialist capacity across facilities~\cite{hu2025human}. This model addresses the WHO-identified bottleneck of specialist availability without requiring that specialists be physically present at every screening site~\cite{who2016recommendations}.

Economically, deploying FADA as a screening support tool requires a one-time computational investment (model training) and minimal ongoing infrastructure compared to the recurring costs of training and retaining additional specialist sonographers, a resource that many LMIC health systems cannot scale sufficiently to meet demand~\cite{kim2017obstetric,edge2025transforming}.

\paragraph{Limitations.}
First, all training data derive from publicly and privately available datasets that may not fully represent the diversity of imaging equipment, patient populations, and pathological conditions encountered in LMIC settings. Second, the normalized coordinate system ([0, 1000) for both detection and segmentation) introduces quantization artifacts for sub-pixel structures such as nuchal translucency membranes (NT Dice: 0.620--0.633), which may require task-specific output resolution in future work. Third, classification performance is highly class-dependent: standard imaging planes achieve $>$95\% accuracy, but fetal pose categories from the FPUS23 dataset remain challenging (5--28\% accuracy), reflecting subtle inter-class visual differences in limb positioning that may exceed current model resolution. Fourth, the interpretation-first pipeline design means that early-stage errors (e.g., incorrect view classification) can propagate through subsequent phases; the human-in-the-loop mode partially mitigates this but autonomous deployment in novel anatomical contexts (e.g., aorta views) remains unreliable. Finally, while the interpretation dataset covers 14 anatomical categories, rare anomalies and pathological presentations are underrepresented, and the 0.8B edge model shows NT segmentation decline (Dice: 0.494), suggesting that extreme compression may require task-specific fine-tuning for challenging structures.


\section{Methods}\label{sec:methods}

\begin{figure}[H]
    \centering
    \includegraphics[width=0.99\textwidth]{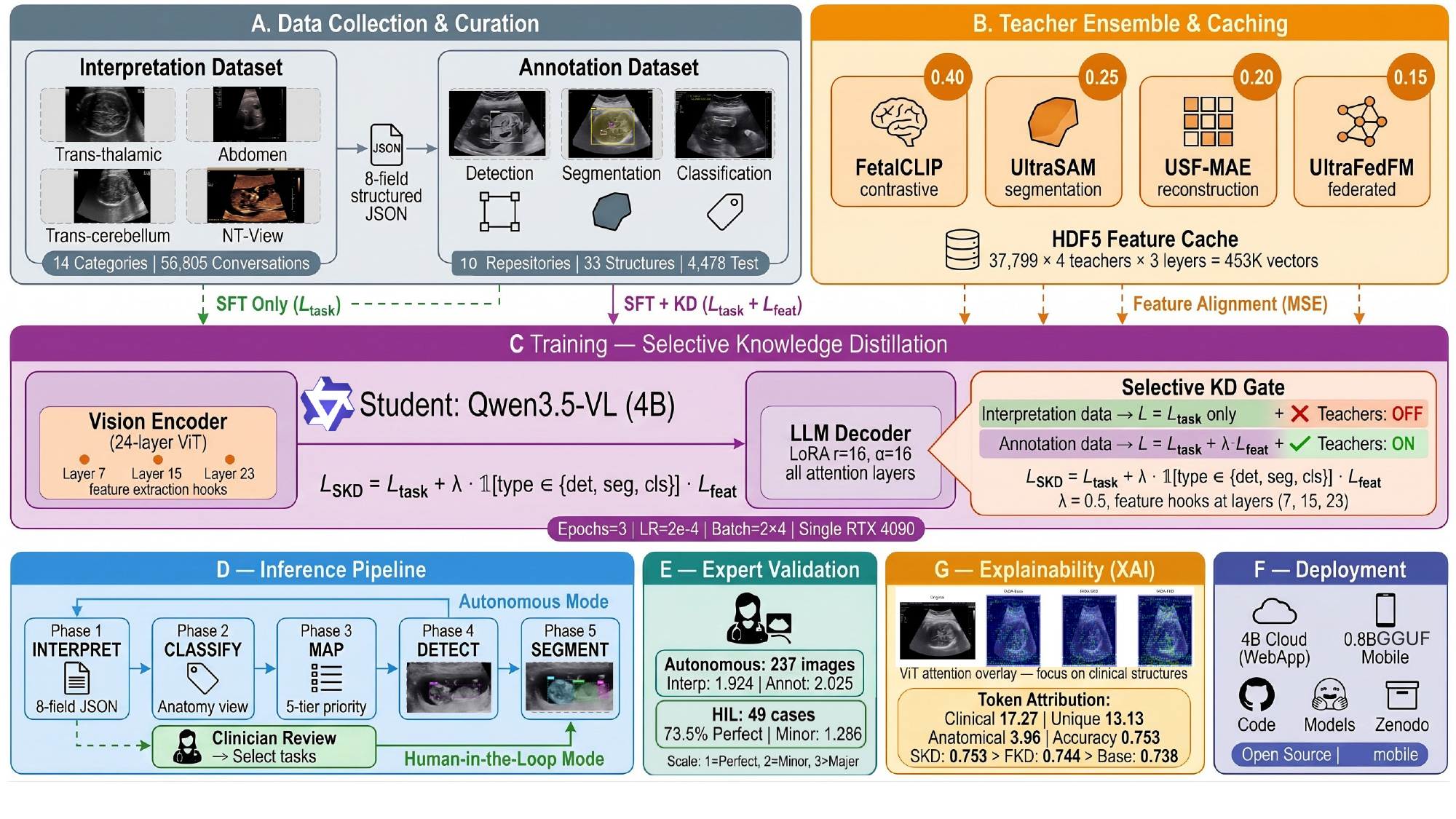}
    \caption{Complete FADA-SKD system lifecycle. (A)~Data collection: 16,478 images spanning 37 views from 8 public and 2 private datasets plus 56,805 interpretation conversations. (B)~Four-teacher ensemble with offline HDF5 feature caching (453K vectors). (C)~Selective Knowledge Distillation: Qwen3.5-VL student with LoRA on a single RTX~4090; feature alignment applied only to annotation data. (D)~5-phase inference pipeline producing detection and segmentation overlays. (E)~Expert validation: autonomous (237 images; interpretation mean 1.924, annotation mean 2.025) and human-in-the-loop (49 cases; 73.5\% perfect interpretations). (F)~Deployment: cloud (4B), web application, and mobile edge via GGUF Q4\_K\_M quantization with llama.cpp (${\sim}$59\,s full pipeline on Android). (G)~Explainability via attention heatmaps and token attribution (field accuracy 0.753).}

    \label{fig:workflow}
\end{figure}

Figure~\ref{fig:workflow} presents the complete FADA pipeline from data collection through deployment and validation.

\subsection{Dataset Curation}\label{sec:dataset}

FADA is trained and evaluated on two complementary datasets curated for this work.

\paragraph{Annotation Dataset.}
Eight publicly available fetal ultrasound datasets and two private collections (Table~\ref{tab:annotation_data}) were aggregated into a unified JSONL format with standardized class names and 8 co-occurrence groups. The combined held-out test set yields 4,478 samples spanning 1,463 detection, 544 segmentation, 2,400 classification, and 71 keypoint instances.

\begin{table}[!t]
\caption{Annotation dataset composition (8 public sources + 2 private).}\label{tab:annotation_data}
\begin{tabular*}{\textwidth}{@{\extracolsep\fill}llll}
\toprule
\textbf{Dataset} & \textbf{Task} & \textbf{Classes} & \textbf{Anatomy} \\
\midrule
Dataset for Fetus Framework~\cite{hussain2022fetus,hussain2022fetus_data} & Detection & 9 & First trimester (nasal, NT) \\
Fast-U-Net~\cite{ashkani2022fast,ashkani2022fast_data} & Segmentation & 2 & Fetal head, abdomen \\
Fetal Abdominal~\cite{fetal_abdominal_mendeley} & Segmentation & 4 & Vessels/organs \\
Fetal Echo FT~\cite{saerens2025fetal_echo} & Classification & 5 & Cardiac views \\
Fetal\_Head~\cite{fetal_head_zenodo} & Segmentation & 3 & Brain (BPD) \\
FOCUS~\cite{focus_dataset} & Detection & 2 & Cardiac (4CH) \\
FPUS23~\cite{fpus23} & Classification & 6 & Fetal pose \\
Pubic Symphysis-FH~\cite{pubic_symphysis_ieee} & Det + Seg & 3 & Pelvis \\
\midrule
CRL\_NT (private) & Det + Seg & 14 & First trimester \\
FUSEP (private) & Detection & 14 & Brain (5 groups) \\
\bottomrule
\end{tabular*}
\end{table}

\paragraph{Interpretation Dataset.}
A total of 56,805 structured clinical conversations (Table~\ref{tab:interp_data}) spanning 14 anatomical categories were curated, enabling cross-task knowledge transfer. Source images were drawn from: AFUSD~\cite{afusd_zenodo} (Abdomen, Aorta, Cervical, Cervix, Femur, Thorax), Fetal\_Head~\cite{fetal_head_zenodo} (Trans-cerebellum, Trans-thalamic, Trans-ventricular), NT dataset~\cite{hussain2022fetus_data} (Standard\_NT, Non\_standard\_NT), Pubic Symphysis~\cite{pubic_symphysis_zenodo} (Public\_Symphysis\_fetal\_head), and two private collections (CRL-View, NT-View).

\begin{table}[!t]
\caption{Interpretation dataset: structured 8-field JSON clinical interpretations per image.}\label{tab:interp_data}
\begin{tabular*}{\textwidth}{@{\extracolsep\fill}lr}
\toprule
\textbf{Property} & \textbf{Value} \\
\midrule
Total conversations & 56,805 \\
Full 8-field format & 37,870 \\
Alternative templates (field subsets) & 18,935 \\
Anatomical categories covered & 14 \\
\midrule
\multicolumn{2}{l}{\textit{Schema fields:}} \\
\multicolumn{2}{l}{\footnotesize anatomical structures, fetal orientation, imaging plane, biometrics,} \\
\multicolumn{2}{l}{\footnotesize gestational age, image quality, normality, clinical recommendations} \\
\bottomrule
\end{tabular*}
\end{table}

\subsection{Model Architecture}\label{sec:architecture}

FADA is built on Qwen3.5-VL~\cite{bai2025qwen25vl}, a vision-language model with a 24-block vision encoder (ViT) producing 1024-dimensional feature vectors and a transformer language decoder. Two scales are evaluated: 4B parameters (primary) and 0.8B (edge deployment). Low-Rank Adaptation (LoRA)~\cite{hu2022lora} is applied with rank $r=16$ and scaling factor $\alpha=16$ to both vision and language attention layers (q\_proj, k\_proj, v\_proj, o\_proj, gate\_proj, up\_proj, down\_proj), yielding approximately 2\% trainable parameters. This enables fine-tuning on a single consumer GPU (24~GB VRAM) while preserving pre-trained generalization.

\subsection{Teacher Models and Feature Pre-computation}\label{sec:teachers}

Four domain-specific ultrasound foundation models serve as knowledge sources, selected for complementary expertise:

\begin{itemize}
    \item \textbf{FetalCLIP}~\cite{fetalclip}: CLIP-based model pre-trained on fetal ultrasound image-text pairs; 24-block ViT-L encoder producing 1024-dimensional features (distillation weight $w=0.4$).
    
    \item \textbf{UltraSAM}~\cite{ultrasam}: Segment Anything Model adapted for ultrasound; 12-block ViT-B encoder producing 768-dimensional spatial features ($w=0.25$).
    
    \item \textbf{USF-MAE}~\cite{usfmae}: Masked autoencoder pre-trained on 43 ultrasound datasets over 500 epochs; 12-block ViT-B producing 768-dimensional features ($w=0.2$).
    
    \item \textbf{UltraFedFM}~\cite{ultrafedfm}: Federated foundation model trained across multiple ultrasound domains; 12-block ViT-B producing 768-dimensional features ($w=0.15$).
\end{itemize}

Teacher features are pre-computed offline using each teacher's vision encoder on the full training set and cached in HDF5 format indexed by image hash. This eliminates concurrent teacher model inference during training, reducing peak GPU memory from $>$80~GB (4 teachers + student) to $<$24~GB (student only + cached features loaded from disk).

\subsection{Offline Knowledge Distillation Framework}\label{sec:kd}

The distillation framework aligns student intermediate features with pre-computed teacher features through learned projector networks. Student features are extracted at layers [7, 15, 23] of the 24-block vision encoder, corresponding to proportional depth matching with teacher architectures (early, mid, and late representations).

For each teacher $t$, a projector network $P_t$ transforms the student feature $\mathbf{h}_s^{(l)}$ at layer $l$ to match the teacher's feature dimensionality:
\begin{equation}
    P_t(\mathbf{h}) = W_2 \cdot \text{GELU}(\text{LayerNorm}(W_1 \cdot \mathbf{h}))
\end{equation}
where $W_1 \in \mathbb{R}^{d_t \times d_s}$ and $W_2 \in \mathbb{R}^{d_t \times d_t}$. The feature alignment loss is computed as:
\begin{equation}
    \mathcal{L}_{\text{feat}} = \sum_{t} w_t \cdot \text{MSE}(P_t(\mathbf{h}_s^{(l_t)}), \mathbf{h}_t)
\end{equation}
where $w_t$ is the teacher importance weight, $l_t$ is the matched student layer, and $\mathbf{h}_t$ is the cached teacher feature. The total training loss combines the task-specific language modeling loss with feature alignment:
\begin{equation}
    \mathcal{L}_{\text{total}} = \mathcal{L}_{\text{task}} + \lambda \cdot \mathcal{L}_{\text{feat}}, \quad \lambda = 0.5
\end{equation}

\subsection{Selective Knowledge Distillation}\label{sec:selective_kd}

The key innovation of FADA-SKD is the conditional application of $\mathcal{L}_{\text{feat}}$ based on data type:
\begin{equation}
    \mathcal{L}_{\text{SKD}} = \mathcal{L}_{\text{task}} + \lambda \cdot \mathbb{1}[\text{type} \in \{\text{det}, \text{seg}, \text{cls}\}] \cdot \mathcal{L}_{\text{feat}}
\end{equation}
where $\mathbb{1}[\cdot]$ is the indicator function. For interpretation data, training proceeds with standard supervised fine-tuning ($\mathcal{L}_{\text{task}}$ only). For annotation data (detection, segmentation, classification), the full distillation loss is applied. This selective strategy is motivated by the observation that teacher models encode visual-spatial patterns optimized for structure localization, which provide complementary supervision for annotation tasks but introduce conflicting optimization objectives for free-text clinical interpretation generation.

In contrast, FADA-FKD applies $\mathcal{L}_{\text{feat}}$ unconditionally to all training batches regardless of data type.

\subsection{Training Protocol}\label{sec:training}

All model variants are trained with identical hyperparameters: learning rate $2 \times 10^{-4}$ with cosine scheduling and 10\% linear warmup, effective batch size of 8 (micro-batch size 2, gradient accumulation over 4 steps), 3 training epochs (42,285 steps), AdamW optimizer with weight decay $10^{-3}$, and bf16 mixed precision. Training is conducted on a single NVIDIA RTX~4090 GPU (24~GB) using the Unsloth~\cite{unsloth} framework for memory-efficient fine-tuning. Each training run requires approximately 40 hours.

\subsection{Interpretation-First Pipeline}\label{sec:pipeline}

At inference, FADA processes each image through a 5-phase cascade where each phase conditions on outputs of preceding phases:

\begin{enumerate}
    \item \textbf{INTERPRET}: Generate an 8-field JSON clinical assessment (anatomical structures, fetal orientation, imaging plane, biometric measurements, gestational age estimation, image quality, normality assessment, clinical recommendations). This serves as the semantic foundation for all downstream tasks.
    
    \item \textbf{CLASSIFY}: Determine the specific anatomical view type (e.g., ``BPD plane'', ``four-chamber view'') from the interpretation output, resolving imaging context for structure mapping.
    
    \item \textbf{MAP}: Apply a 5-tier priority cascade to determine appropriate detection and segmentation targets: (1)~specific label match from interpretation text, (2)~imaging plane match to co-occurrence group, (3)~keyword scoring with field-weighted matching, (4)~generic label fallback, (5)~default anatomical classes. This replaces external class labels entirely.
    
    \item \textbf{DETECT}: Execute targeted bounding-box detection using the mapped class set, producing normalized coordinates in [0,\,1000) format.
    
    \item \textbf{SEGMENT}: Execute targeted polygon segmentation using the mapped class set, producing vertex sequences in the same normalized coordinate space.
\end{enumerate}

\paragraph{Autonomous Mode.}
In fully autonomous mode, all five phases execute sequentially without human intervention. The interpretation output propagates through classification and mapping to determine detection/segmentation targets automatically, enabling deployment by non-specialist operators who simply upload an image and receive complete structured analysis.

\paragraph{Human-in-the-Loop Mode.}
In HiL mode, a clinician reviews the Phase~1 interpretation before subsequent phases execute. The operator can: (a)~accept the interpretation and proceed with full pipeline execution, (b)~override the classified view type to correct mapping errors, (c)~selectively execute only specific phases (e.g., segmentation without detection), or (d)~specify target structures directly, bypassing the MAP phase. This interactive design mitigates error propagation from early-stage misinterpretation, the primary failure mode identified in autonomous deployment (Supplementary Table~S4).

This pipeline eliminates the need for external class labels at inference. The model autonomously determines what to detect and segment based on its own clinical interpretation, making it suitable for deployment without sonographer expertise.

\subsection{Evaluation Protocol}\label{sec:eval_protocol}

\paragraph{Automated Metrics.}
Detection is evaluated using mean Average Precision at IoU thresholds 0.50, 0.75, and 0.50:0.95 across 33 structure classes. Segmentation uses Dice coefficient and IoU across 10 structure classes with polygon-to-mask conversion. Classification accuracy is computed with exact string matching.

\paragraph{Expert Evaluation.}
An expert sonographer with $>$10 years of clinical experience evaluated 237 images across 18 anatomical categories. For each image, all three model outputs were presented in randomized order without model identification. The sonographer scored each output from 1 (clinically acceptable) to 3 (poor quality) for annotation accuracy and interpretation correctness independently. A subset of 49 cases was additionally evaluated in human-in-the-loop mode using the deployed web application.


\section{Conclusion}\label{sec:conclusion}

FADA is a unified vision-language model for fetal ultrasound analysis that combines clinical interpretation, detection, and segmentation within a single architecture through an interpretation-first pipeline. The central finding is that selective knowledge distillation, applying feature alignment from domain-specific teachers only to annotation data, outperforms full distillation: achieving the best segmentation performance (Dice: 0.8820), the best mAP@0.75 among SKD/Base comparison (0.4402), and the highest expert-rated interpretation quality (mean score: 1.924) among all evaluated variants. This finding carries broader implications for multi-task VLM training: auxiliary losses should be selectively applied based on task compatibility rather than uniformly across all training data.

Human-in-the-loop evaluation demonstrates that FADA-SKD achieves 73.5\% perfect interpretation scores when deployed with minimal operator guidance, indicating viability as a clinical decision support tool in resource-constrained settings. The open-source web application and compressed 0.8B model variant provide deployment pathways for both connected and offline clinical environments. Critically, we validate edge deployment by running the full 5-phase pipeline entirely on a commodity smartphone (Snapdragon~7 Gen~1, 12\,GB RAM) in approximately 59\,s without network connectivity, demonstrating that the model can be integrated with fetal ultrasound devices in a stand-alone fashion for point-of-care screening.

Future work will focus on expanding training data to include pathological presentations and rare anomalies, multi-language interpretation generation for diverse clinical settings, prospective clinical validation studies in LMIC facilities, and further optimization of on-device inference latency through hardware-specific quantization and speculative decoding techniques.

\subsection{Use of AI-Assisted Tools}\label{sec:ai_disclosure}
A large language model was used during manuscript preparation to edit, revise, and improve the clarity of written text. AI-assisted tools were also used to design the visual layout and arrangement of the workflow diagram (Figure~\ref{fig:workflow}); however, all images within the diagram panels are authentic and were taken directly from the study materials and placed manually by the authors. No scientific content, including experimental results, clinical interpretations, or quantitative analyses, was generated by AI tools. The authors reviewed all AI-assisted outputs and take full responsibility for the accuracy and integrity of the published work.


\backmatter

\bmhead{Acknowledgments}

This work was funded by the Canadian International Development Research Centre (IDRC) under Grant Agreement No. 110060-001, managed by the Global Health Institute at the American University of Beirut through the Global Health and Artificial Intelligence Network in MENA (GHAIN MENA). This study forms part of a broader research program on responsible AI for development. This publication was also supported by the PPM 7th Cycle grant (PPM 07-0409-240041, AMAL-For-Qatar) from the Qatar Research, Development, and Innovation Council (QRDI Council), a member of Qatar Foundation. The authors also thank Dr. Shalal Mohsen for his valuable contribution as an expert sonographer in the evaluation of the proposed system. The findings and conclusions presented in this publication are solely the responsibility of the authors.

\section*{Declarations}

\bmhead{Funding}
Canadian International Development Research Centre (IDRC), Grant 110060-001. Qatar Research Development and Innovation Council (QRDI), Grant PPM 07-0409-240041.

\bmhead{Ethics approval}
This study uses publicly available de-identified ultrasound datasets. Expert sonographer evaluation constitutes professional consultation and does not require separate IRB approval. No patient-identifiable data were collected or used.

\bmhead{Data availability}
The interpretation training dataset (56,805 structured clinical conversations with expert sonographer annotations) is available on Zenodo (DOI: \texttt{https://doi.org/10.5281/zenodo.20381238}) under a CC-BY-4.0 license. Source ultrasound images for the annotation dataset originate from publicly available repositories: Dataset for Fetus Framework~\cite{hussain2022fetus_data}, Fast-U-Net~\cite{ashkani2022fast_data}, Fetal Abdominal Structures~\cite{fetal_abdominal_mendeley}, Fetal Echocardiography First Trimester~\cite{saerens2025fetal_echo}, Fetal\_Head~\cite{fetal_head_zenodo}, FOCUS~\cite{focus_dataset}, FPUS23~\cite{fpus23}, and Pubic Symphysis-Fetal Head~\cite{pubic_symphysis_ieee}. Interpretation dataset images derive from AFUSD~\cite{afusd_zenodo}, Fetal\_Head~\cite{fetal_head_zenodo}, NT dataset~\cite{hussain2022fetus_data}, and Pubic Symphysis~\cite{pubic_symphysis_zenodo}. CRL\_NT and FUSEP remain private due to institutional restrictions. Evaluation data and sonographer scoring sheets are included in the supplementary materials.

\bmhead{Code availability}
All code for model training, evaluation, and the web application is available at \url{https://github.com/mahmoodphd/FADA} under the Apache 2.0 license. Model weights are available on HuggingFace Hub (\url{https://huggingface.co/mshz88/FADA-SKD-4B}). The GGUF-quantized mobile model for on-device inference via llama.cpp is available at \url{https://huggingface.co/mshz88/FADA-Mobile-GGUF}. The Android application (APK) is available for download from the GitHub repository releases. The interactive web application is deployed at \url{https://mshz88-fada-ultrasound-vlm.hf.space.}. Video demonstrations of the autonomous and human-in-the-loop deployment modes are available at YouTube at \url{https://youtu.be/CbXcz74fn6k} and offline mobile app demo at \url{https://youtu.be/RoogJqPNZ4w}.

\bmhead{Conflict of interest}
The authors declare no competing interests.

\bmhead{Author contributions}
M.A. conceived the study, designed the system architecture, developed the training and inference pipelines, implemented the selective knowledge distillation framework, built the web application and mobile Android app, conducted all computational experiments, and wrote the manuscript. U.S. assisted with data preprocessing and annotation pipeline development. R.M. contributed to evaluation scripting and result analysis. I.A. contributed to literature review and manuscript editing. N.M. provided clinical guidance on fetal ultrasound interpretation standards and validated anatomical correctness of model outputs. A.M. performed expert sonographer evaluation across all 237 autonomous images and 49 human-in-the-loop cases. K.A. provided clinical oversight and reviewed the clinical relevance of system outputs. M.H. provided project supervision, and reviewed the manuscript. M.Ag. supervised the research, guided the experimental design, and critically revised the manuscript. All authors read and approved the final manuscript.


\bibliography{sn-bibliography}

\end{document}


\maketitle

\section{Dataset Sources and Provenance}

Tables~\ref{tab:annotation_sources} and~\ref{tab:interpretation_sources} enumerate the public and private data sources used to construct the FADA annotation and interpretation training datasets. Most public datasets are released under open licenses (CC-BY-4.0 or equivalent).
\begin{table*}[htbp]
\centering
\footnotesize
\caption{Annotation dataset sources. Each row corresponds to a source dataset used for detection, segmentation, classification, or keypoint annotation training and evaluation.}
\label{tab:annotation_sources}

\begin{tabularx}{\textwidth}{
>{\raggedright\arraybackslash}p{3.2cm}
>{\raggedright\arraybackslash}p{2cm}
>{\raggedright\arraybackslash}p{3cm}
X
}
\toprule
\textbf{Dataset} & \textbf{Task} & \textbf{Source} & \textbf{Access} \\
\midrule

Dataset for Fetus Framework &
Detection &
Hussain et al.\ (2022) &
\url{https://data.mendeley.com/datasets/yrzzw9m6kk/1} \\

Fast-U-Net &
Segmentation &
Ashkani et al.\ (2022) &
\url{https://github.com/vahidashkani/Fast-U-Net} \\

Fetal Abdominal Structures &
Segmentation &
Mendeley Data (2023) &
\url{https://data.mendeley.com/datasets/4gcpm9dsc3/1} \\

Fetal Echo First Trimester &
Classification &
Saerens et al.\ (2025) &
\url{https://www.nature.com/articles/s41746-025-02217-6} \\

Fetal\_Head &
Segmentation &
Zenodo (2023) &
\url{https://zenodo.org/records/8265464} \\

FOCUS Dataset &
Detection &
Zenodo (2025) &
\url{https://zenodo.org/records/14597550} \\

FPUS23 &
Classification &
Prabakaran et al.\ (2023) &
\url{https://ieeexplore.ieee.org/document/10146252} \\

Pubic Symphysis-FH &
Det + Seg &
Lu et al.\ (2024) &
\url{https://ieeexplore.ieee.org/document/10529289} \\

\midrule

CRL\_NT &
Det + Seg &
Private &
--- \\

FUSEP &
Detection &
Private &
--- \\

\bottomrule
\end{tabularx}
\end{table*}
\begin{table*}[htbp]
\centering
\footnotesize
\caption{Interpretation dataset sources. Each row lists the origin of ultrasound images used to generate the 56,805 structured clinical interpretation conversations.}
\label{tab:interpretation_sources}

\begin{tabularx}{\textwidth}{
>{\raggedright\arraybackslash}p{5cm}
>{\raggedright\arraybackslash}p{3cm}
X
}
\toprule
\textbf{Categories} & \textbf{Source} & \textbf{Access} \\
\midrule

Abdomen, Aorta, Cervical, Cervix, Femur, Thorax &
AFUSD (2020) &
\url{https://zenodo.org/records/3904280} \\

Trans-cerebellum, Trans-thalamic, Trans-ventricular &
Fetal\_Head (2023) &
\url{https://zenodo.org/records/8265464} \\

Standard\_NT, Non\_standard\_NT &
NT Dataset (2023) &
\url{https://data.mendeley.com/datasets/n2rbrb9t4f/1} \\

Public\_Symphysis\_fetal\_head &
Pubic Symphysis (2024) &
\url{https://zenodo.org/records/10969427} \\

\midrule

CRL-View, NT-View &
Private &
--- \\

\bottomrule
\end{tabularx}
\end{table*}
\section{Table S1: Per-Class Detection Performance}
\footnotesize
\begin{longtable}{lccccc}
\caption{Per-class detection AP@0.50 across FADA model variants. Best result per class in \textbf{bold}, second-best \underline{underlined}. Models: Base = SFT only (no KD), SKD = selective KD (annotation data only), FKD = full KD (all data). Class abbreviations: AB=Abdomen, B=Body, C=Chest, CSP=Cavum Septum Pellucidum, CRL=Crown-Rump Length, DP=Diaphragm, G=Genitalia, H=Head, LV=Lateral Ventricle, MDS=Midline Septum, MLS=Midline Structures, MX=Maxilla, NB=Nasal Bone, NT=Nuchal Translucency, NTAPS=NT Attachment Points, RBP=Rib/Body Profile.}
\label{tab:perclass_detection}\\
\toprule
Class & FADA-Base & FADA-SKD & FADA-FKD & FADA-Base & FADA-SKD \\
      & (4B) & (4B) & (4B) & (0.8B) & (0.8B) \\
\midrule
\endfirsthead
\multicolumn{6}{c}{\textit{Table S1 continued from previous page}}\\
\toprule
Class & FADA-Base & FADA-SKD & FADA-FKD & FADA-Base & FADA-SKD \\
      & (4B) & (4B) & (4B) & (0.8B) & (0.8B) \\
\midrule
\endhead
\midrule
\multicolumn{6}{r}{\textit{Continued on next page}}\\
\endfoot
\bottomrule
\endlastfoot
AB              & \best{0.745} & 0.665 & \second{0.698} & 0.670 & 0.695 \\
B               & 0.815 & 0.796 & \best{0.813} & 0.768 & 0.799 \\
Brain           & \best{1.000} & \best{1.000} & \best{1.000} & \best{1.000} & \best{1.000} \\
C               & \best{0.781} & 0.682 & \second{0.732} & 0.718 & 0.717 \\
CRL             & \best{1.000} & \best{1.000} & \best{1.000} & \best{1.000} & \best{1.000} \\
CSP             & \second{0.822} & 0.809 & \best{0.841} & 0.727 & 0.674 \\
DP              & \best{0.770} & 0.701 & \second{0.724} & 0.698 & 0.693 \\
G               & \best{0.643} & \second{0.589} & 0.559 & 0.496 & 0.457 \\
H               & \best{0.872} & 0.808 & \second{0.833} & 0.830 & 0.827 \\
LV              & \best{0.683} & \second{0.647} & 0.701 & 0.499 & 0.532 \\
MDS             & \best{0.625} & 0.586 & \second{0.605} & 0.457 & 0.425 \\
MLS             & \best{0.707} & \second{0.655} & \best{0.709} & 0.362 & 0.414 \\
MX              & \best{0.750} & \second{0.678} & 0.726 & 0.600 & 0.572 \\
NB              & 0.342 & \best{0.391} & \second{0.371} & 0.241 & 0.194 \\
NT              & \best{0.662} & 0.578 & \second{0.638} & 0.562 & 0.534 \\
NTAPS           & 0.591 & \second{0.603} & \best{0.609} & 0.435 & 0.313 \\
RBP             & \best{0.804} & 0.692 & \second{0.781} & 0.662 & 0.657 \\
ScaleBar        & 0.802 & \best{1.000} & \second{0.898} & 0.725 & 0.588 \\
ScaleBarKpoints & 0.802 & \best{1.000} & \second{0.898} & 0.725 & 0.588 \\
abdomen         & \best{0.987} & \best{0.990} & \best{0.988} & \best{0.987} & \best{0.987} \\
arm             & 0.925 & 0.913 & \best{0.936} & \best{0.931} & \best{0.932} \\
artery          & \best{0.574} & \second{0.551} & 0.512 & 0.496 & 0.481 \\
cardiac         & \best{1.000} & \best{1.000} & \best{1.000} & \best{1.000} & \best{1.000} \\
fetal\_head     & \best{1.000} & \best{1.000} & \best{1.000} & \best{1.000} & \best{1.000} \\
head            & \best{1.000} & \best{1.000} & \best{1.000} & \best{1.000} & 0.984 \\
legs            & 0.858 & \best{0.920} & \second{0.915} & \second{0.869} & \best{0.894} \\
liver           & 0.964 & \best{0.970} & \second{0.970} & 0.963 & \best{0.975} \\
nasal\_bone     & 0.442 & \best{0.519} & \second{0.526} & 0.250 & 0.163 \\
nasal\_skin     & \best{0.357} & \second{0.193} & 0.219 & 0.122 & 0.091 \\
nasal\_tip      & \best{0.671} & \second{0.668} & 0.582 & 0.240 & 0.347 \\
pubic\_symphysis & 0.988 & 0.990 & \best{0.989} & \best{1.000} & \best{1.000} \\
stomach         & \best{0.844} & \second{0.821} & 0.739 & 0.775 & 0.764 \\
thorax          & \best{1.000} & \best{1.000} & \best{1.000} & \best{1.000} & \best{1.000} \\
vein            & \best{0.690} & \second{0.667} & 0.651 & 0.602 & 0.633 \\
\midrule
\textbf{mAP@0.50} & \best{0.780} & \second{0.767} & 0.770 & 0.689 & 0.674 \\
\textbf{mAP@0.75} & 0.421 & \second{0.440} & \best{0.458} & 0.382 & 0.376 \\
\end{longtable}
\normalsize

\noindent FADA-Base (4B) leads overall mAP@0.50 (0.780), followed by FADA-FKD (0.770) and FADA-SKD (0.767). FADA-FKD (4B) achieves the best mAP@0.75 (0.458), consistent with tighter bounding-box predictions. FADA-SKD (4B) surpasses Base on fine-grained structures (ScaleBar, nasal\_bone, legs) while slightly trailing on several FUSEP classes (AB, C, MX). The 0.8B models retain 85--88\% of 4B performance despite 5$\times$ fewer parameters.

\section{Table S2: Per-Dataset Performance Breakdown}

\begin{table}[htbp]
\centering
\footnotesize
\caption{Detection performance (mAP@0.50) by source dataset across model variants.}
\label{tab:perdataset_detection}
\begin{tabular}{lccc}
\toprule
Dataset & FADA-Base & FADA-SKD & FADA-FKD \\
        & (4B) & (4B) & (4B) \\
\midrule
FOCUS Dataset (cardiac/thorax) & \best{1.000} & \best{1.000} & \best{1.000} \\
Pubic Symphysis (\_slice\_cache) & 0.994 & \best{0.995} & \best{0.995} \\
FPUS23 (abdomen/arm/head/legs) & 0.954 & \second{0.955} & \best{0.960} \\
Fetal\_Head (Brain/CSP/LV) & \second{0.801} & 0.790 & \best{0.835} \\
CRL\_NT (first trimester) & \best{0.750} & \second{0.737} & 0.738 \\
Fetal Abdominal Structures & \best{0.768} & \second{0.730} & 0.718 \\
FUSEP (multi-structure) & \best{0.695} & \second{0.672} & 0.649 \\
\bottomrule
\end{tabular}
\end{table}

\begin{table}[htbp]
\centering
\footnotesize
\caption{Segmentation performance (Mean Dice) by source dataset.}
\label{tab:perdataset_segmentation}
\begin{tabular}{lcccc}
\toprule
Dataset & FADA-Base & FADA-SKD & FADA-FKD & FADA-Base \\
        & (4B) & (4B) & (4B) & (0.8B) \\
\midrule
Fetal\_Head (Brain) & 0.969 & \second{0.970} & \best{0.971} & 0.966 \\
Pubic Symphysis (\_slice\_cache) & 0.952 & \best{0.955} & \second{0.956} & 0.946 \\
FOCUS Dataset (cardiac/thorax) & \best{0.928} & \second{0.921} & 0.925 & 0.919 \\
Fetal Abdominal Structures & 0.745 & \best{0.750} & 0.737 & 0.700 \\
CRL\_NT (NT only) & \best{0.633} & \second{0.624} & 0.620 & 0.651 \\
\midrule
\textbf{Global Mean Dice} & 0.881 & \best{0.882} & 0.879 & 0.863 \\
\textbf{Global Mean IoU} & 0.813 & \best{0.815} & 0.811 & 0.790 \\
\bottomrule
\end{tabular}
\end{table}

\begin{table}[htbp]
\footnotesize
\centering
\caption{Classification accuracy by source dataset.}
\label{tab:perdataset_classification}
\begin{tabular}{lccc}
\toprule
Dataset & FADA-Base & FADA-SKD & FADA-FKD \\
        & (4B) & (4B) & (4B) \\
\midrule
Fetal Echocardiography First Trimester & 0.904 & \best{0.907} & 0.893 \\
FPUS23 Dataset & 0.713 & \best{0.744} & \second{0.743} \\
\midrule
\textbf{Overall Accuracy} & 0.823 & \best{0.838} & \second{0.830} \\
\bottomrule
\end{tabular}
\end{table}

\noindent SKD improves segmentation across most datasets with negligible detection loss. FKD achieves competitive detection (mAP@0.50=0.770) with improved fine-grained localization (best mAP@0.75) but slightly lower classification accuracy than SKD. The FOCUS and Pubic Symphysis datasets are saturated ($\approx$1.0) across all models. The largest inter-model variance concentrates on fine-grained structures (CRL\_NT, Fetal Abdominal), where differences in distillation strategy are most apparent.

\section{Table S3: Sonographer Assessment Summary}

\subsection{Scoring Protocol}
An experienced sonographer independently evaluated 237 ultrasound images (62 external clinical images and 175 from the held-out test set) across all three FADA 4B model variants. Both annotation quality and interpretation quality were scored on a 3-point scale:
\begin{itemize}
    \item \textbf{Score 1} (Correct): Output is clinically acceptable
    \item \textbf{Score 2} (Partial): Output contains minor errors but retains partial clinical utility
    \item \textbf{Score 3} (Failure): Output is clinically unacceptable
\end{itemize}

\begin{table}[htbp]
\centering
\caption{Sonographer assessment score distribution by model (n=237 images). Lower scores indicate better performance.}
\label{tab:sonographer_scores}
\footnotesize
\begin{tabular}{llccc|c}
\toprule
Task & Model & Score=1 & Score=2 & Score=3 & Mean Score \\
     &       & (Correct) & (Partial) & (Failure) & \\
\midrule
\multirow{3}{*}{Annotation}
 & FADA-Base (4B) & 86 (36.3\%) & 61 (25.7\%) & 90 (38.0\%) & 2.017 \\
 & FADA-SKD (4B)  & 83 (35.0\%) & 65 (27.4\%) & 89 (37.6\%) & 2.025 \\
 & FADA-FKD (4B)  & 79 (33.3\%) & 67 (28.3\%) & 91 (38.4\%) & 2.051 \\
\midrule
\multirow{3}{*}{Interpretation}
 & FADA-Base (4B) & 70 (29.5\%) & 71 (30.0\%) & 96 (40.5\%) & 2.110 \\
 & FADA-SKD (4B)  & 90 (38.0\%) & 75 (31.6\%) & 72 (30.4\%) & \best{1.924} \\
 & FADA-FKD (4B)  & 65 (27.4\%) & 64 (27.0\%) & 108 (45.6\%) & 2.181 \\
\bottomrule
\end{tabular}
\end{table}

\begin{table}[htbp]
\centering
\footnotesize
\caption{Failure mode distribution across models from sonographer assessment.}
\label{tab:model_failure_comparison}
\begin{tabular}{lccc}
\toprule
Model & Annotation Failures & Interpretation Failures & Total Failures \\
      & (Score $\geq$ 2) & (Score $\geq$ 2) & \\
\midrule
FADA-Base (4B) & 151 & 167 & 318 \\
FADA-SKD (4B) & 154 & \best{147} & \best{301} \\
FADA-FKD (4B) & 158 & 172 & 330 \\
\bottomrule
\end{tabular}
\end{table}

\noindent FADA-SKD (4B) achieves the strongest interpretation quality (mean score 1.924 vs 2.110 for Base), consistent with selective KD preserving the student model's language generation capabilities while maintaining comparable annotation performance.

\section{Table S4: Failure Mode Catalog}

\begin{table}[htbp]
\centering
\caption{Failure mode catalog by anatomical structure (aggregated across all 3 models, n=237 images $\times$ 3 models $\times$ 2 tasks = 1,422 assessments).}
\label{tab:failure_modes_anatomy}
\footnotesize
\begin{tabular}{lcccc}

\toprule
Anatomical Structure & Total Failures & Partial (Score=2) & Complete (Score=3) & Failure Rate (\%) \\
\midrule
Brain (Head) & 217 & 122 & 95 & 15.3 \\
Other/Non-standard & 210 & 72 & 138 & 14.8 \\
Limbs (Femur/Arms/Legs) & 158 & 59 & 99 & 11.1 \\
Cardiac & 148 & 60 & 88 & 10.4 \\
Cervix/Pubic Symphysis & 123 & 56 & 67 & 8.6 \\
Abdomen & 110 & 55 & 55 & 7.7 \\
First Trimester (CRL/NT) & 9 & 3 & 6 & 0.6 \\
\midrule
\textbf{Total} & \textbf{975} & \textbf{427} & \textbf{548} & \textbf{68.6} \\
\bottomrule
\end{tabular}
\end{table}

\subsection{Representative Failure Patterns}
\begin{enumerate}
    \item \textbf{Out-of-distribution anatomy} (Severity: Complete failure): Aorta views and symphysis pubic images from external clinical sources absent from the training distribution cause total annotation failure across all models.
    \item \textbf{Fine structure segmentation} (Severity: Partial to Complete): CSP (Cavum Septum Pellucidum) and LV (Lateral Ventricle) segmentation frequently fails or yields inaccurate masks, particularly in non-standard views.
    \item \textbf{View misclassification} (Severity: Partial): Cerebellar views are occasionally interpreted as trans-thalamic; ventricle views are misclassified as sagittal.
    \item \textbf{Prompt sensitivity} (Severity: Partial): Detection accuracy varies with prompt phrasing (e.g., ``detect abdomen'' vs ``detect abdominal structures'' produces different results).
    \item \textbf{Femoral length detection} (Severity: Complete for external): External femur images return leg bounding boxes rather than femur-specific detections.
\end{enumerate}

\section{Table S5: Statistical Significance Tests}

\begin{table}[htbp]
\centering
\footnotesize
\caption{Model performance with 95\% confidence intervals (n=4,478 test samples, 1,000 bootstrap iterations).}
\label{tab:confidence_intervals}
\begin{tabular}{lccccc}
\toprule
Model & mAP@0.50 & mAP@0.75 & Mean Dice & Mean IoU & Cls. Accuracy \\
\midrule
FADA-Base (4B) & 0.780$\pm$0.022 & 0.421$\pm$0.024 & 0.881$\pm$0.028 & 0.813$\pm$0.032 & 0.823$\pm$0.015 \\
FADA-SKD (4B) & 0.767$\pm$0.022 & 0.440$\pm$0.026 & \best{0.882}$\pm$0.026 & \best{0.815}$\pm$0.031 & \best{0.838}$\pm$0.015 \\
FADA-FKD (4B) & 0.770$\pm$0.022 & \best{0.458}$\pm$0.026 & 0.879$\pm$0.026 & 0.811$\pm$0.033 & 0.830$\pm$0.015 \\
\bottomrule
\end{tabular}
\end{table}

\begin{table}[htbp]
\footnotesize
\centering
\caption{Pairwise statistical comparisons (two-proportion z-test). $^{*}$p$<$0.05, ns = not significant.}
\label{tab:pairwise_stats}
\begin{tabular}{lccccl}
\toprule
Comparison & $\Delta$mAP@0.50 & $\Delta$mAP@0.75 & $\Delta$Dice & $\Delta$Cls Acc & Significance \\
\midrule
FADA-SKD (4B) vs Base & $-$0.013 & +0.019 & +0.001 & +0.015 & All ns \\
               & (p=0.41) & (p=0.30) & (p=0.97) & (p=0.16) & \\
\midrule
FADA-SKD (4B) vs FKD & $-$0.003 & $-$0.018 & +0.003 & +0.008 & All ns \\
              & (p=0.85) & (p=0.35) & (p=0.88) & (p=0.45) & \\
\midrule
FKD vs Base & $-$0.010 & +0.037 & $-$0.002 & +0.007 & All ns \\
            & (p=0.53) & (p=0.05) & (p=0.94) & (p=0.51) & \\
\bottomrule
\end{tabular}
\end{table}

\noindent No statistically significant differences emerge between any model pair in detection or segmentation; neither selective nor full KD compromises annotation performance relative to the Base model. FADA-SKD attains the best segmentation (Dice 0.882) and classification accuracy (0.838) while maintaining detection within the confidence interval of Base. FADA-FKD achieves the best mAP@0.75 (0.458), pointing to tighter bounding-box precision.

\section{Table S6: Teacher Contribution Analysis}

\begin{table}[htbp]
\centering
\footnotesize
\caption{Effective teacher contribution by task type. Contribution = KD weight ($\alpha_i$) $\times$ task relevance, normalized per task. Dominant teacher shown in \textbf{bold}.}
\label{tab:teacher_task}
\begin{tabular}{lccccc}
\toprule
Task Type & FetalCLIP & UltraSAM & USF-MAE & UltraFedFM & Dominant \\
          & ($\alpha$=0.40) & ($\alpha$=0.25) & ($\alpha$=0.20) & ($\alpha$=0.15) & \\
\midrule
View Classification & \best{0.672} & 0.084 & 0.168 & 0.076 & FetalCLIP \\
Object Detection & \best{0.505} & 0.270 & 0.144 & 0.081 & FetalCLIP \\
Semantic Segmentation & 0.242 & \best{0.505} & 0.162 & 0.091 & UltraSAM \\
Clinical Interpretation & \best{0.593} & 0.046 & 0.222 & 0.139 & FetalCLIP \\
Keypoint Detection & \best{0.385} & 0.337 & 0.192 & 0.087 & FetalCLIP \\
\bottomrule
\end{tabular}
\end{table}

\begin{table}[htbp]
\centering
\footnotesize
\caption{Teacher model properties and KD configuration.}
\label{tab:teacher_properties}
\begin{tabular}{lccll}
\toprule
Teacher & KD Weight & Feature Dim & Specialization & Pre-training \\
\midrule
FetalCLIP & 0.40 & 1024 & Fetal US classification & Contrastive (VL-aligned) \\
UltraSAM & 0.25 & 768 & US segmentation & SAM-based (spatial) \\
USF-MAE & 0.20 & 768 & General US features & Self-supervised MAE \\
UltraFedFM & 0.15 & 768 & Cross-institution & Federated SSL (19 organs) \\
\bottomrule
\end{tabular}
\end{table}

\begin{table}[htbp]
\centering
\footnotesize
\caption{Teacher importance by anatomical structure (normalized effective contribution).}
\label{tab:teacher_anatomy}
\begin{tabular}{lcccc}
\toprule
Structure & FetalCLIP & UltraSAM & USF-MAE & UltraFedFM \\
\midrule
Brain (Trans-thalamic) & \best{0.45} & 0.20 & 0.20 & 0.15 \\
Brain (Trans-cerebellar) & \best{0.40} & 0.25 & 0.20 & 0.15 \\
Cardiac (4-Chamber) & \best{0.35} & 0.30 & 0.20 & 0.15 \\
Cardiac (Outflow) & 0.30 & \best{0.30} & 0.25 & 0.15 \\
Abdomen & \best{0.35} & 0.30 & 0.20 & 0.15 \\
Femur Length & 0.25 & \best{0.40} & 0.20 & 0.15 \\
NT Measurement & 0.30 & \best{0.35} & 0.20 & 0.15 \\
Pubic Symphysis & 0.25 & \best{0.35} & 0.20 & 0.20 \\
\bottomrule
\end{tabular}
\end{table}

\noindent FetalCLIP dominates classification and interpretation tasks (40--67\% effective weight), a consequence of its contrastive VL pre-training on fetal images. UltraSAM dominates segmentation (50\%) and spatial measurement tasks (femur/NT). This pattern explains why selective KD preserves interpretation quality: teacher features are applied only to annotation data, leaving the student's language generation pathway undistorted by spatial teacher signals.

\section{Table S7: Cross-Task Consistency Analysis}

\begin{table}[htbp]
\centering
\footnotesize
\caption{Cross-task performance consistency. Higher Pearson $r$ indicates stronger detection--segmentation coupling.}
\label{tab:consistency_pearson}
\begin{tabular}{lccccc|c}
\toprule
Model & mAP@0.50 & mAP@0.75 & Mean Dice & Mean IoU & Cls Acc & Det--Seg $r$ \\
\midrule
FADA-Base (4B) & 0.780 & 0.421 & 0.881 & 0.813 & 0.823 & 0.548 \\
FADA-SKD (4B) & 0.767 & 0.440 & \best{0.882} & \best{0.815} & \best{0.838} & \second{0.612} \\
FADA-FKD (4B) & 0.770 & \best{0.458} & 0.879 & 0.811 & 0.830 & \best{0.739} \\
\bottomrule
\end{tabular}
\end{table}

\begin{table}[htbp]
\centering
\footnotesize
\caption{Per-dataset detection--segmentation consistency (gap = |mAP@0.50 $-$ Dice|).}
\label{tab:consistency_gap}
\begin{tabular}{llccc}
\toprule
Model & Dataset & Det mAP@0.50 & Seg Dice & Gap \\
\midrule
\multirow{4}{*}{FADA-Base (4B)}
 & FOCUS Dataset & 1.000 & 0.928 & 0.072 \\
 & Pubic Symphysis & 0.994 & 0.955 & 0.039 \\
 & Fetal\_Head & 0.798 & 0.969 & 0.172 \\
 & Fetal Abdominal & 0.768 & 0.745 & 0.023 \\
\midrule
\multirow{4}{*}{FADA-SKD (4B)}
 & FOCUS Dataset & 1.000 & 0.930 & 0.070 \\
 & Pubic Symphysis & 0.995 & 0.957 & 0.038 \\
 & Fetal\_Head & 0.805 & 0.970 & 0.165 \\
 & Fetal Abdominal & 0.752 & 0.745 & 0.008 \\
\midrule
\multirow{4}{*}{FADA-FKD (4B)}
 & FOCUS Dataset & 1.000 & 0.925 & 0.075 \\
 & Pubic Symphysis & 0.995 & 0.956 & 0.039 \\
 & Fetal\_Head & 0.835 & 0.971 & 0.136 \\
 & Fetal Abdominal & 0.718 & 0.737 & 0.019 \\
\bottomrule
\end{tabular}
\end{table}

\noindent FADA-FKD (4B) achieves the highest detection--segmentation correlation ($r$=0.739 vs 0.612 for SKD and 0.548 for Base), consistent with full knowledge distillation strengthening cross-task spatial coherence. FADA-SKD nonetheless remains the recommended deployment variant owing to superior segmentation, classification, and expert-rated interpretation quality. The largest gap (0.136--0.172) arises on Fetal\_Head, where segmentation (Brain Dice=0.97) substantially outperforms detection (CSP/LV remain challenging). FADA-FKD narrows this gap most effectively (0.136).

\newpage
\section{Table S8: Representative Cases from Sonographer Evaluation}

The following cases from external validation (scored by an experienced sonographer) illustrate both strengths and characteristic failure modes of the FADA system.

\begin{longtable}{p{0.5cm}p{3.5cm}p{1cm}p{1cm}p{6cm}}
\caption{Selected representative cases from sonographer evaluation (external\_validation\_scoring\_v2.csv). I=Interpretation score, A=Annotation score (1=correct, 2=partial, 3=failure).}
\label{tab:representative_cases}\\
\toprule
\# & Image Category & I & A & Sonographer Notes \\
\midrule
\endfirsthead
\toprule
\# & Image Category & I & A & Sonographer Notes \\
\midrule
\endhead
\bottomrule
\endlastfoot

\multicolumn{5}{l}{\textit{\textbf{Success Cases}}} \\
\midrule
1 & Trans-thalamic (test) & 1 & 1 & (Correct detection and interpretation) \\
5 & Cardiac aorta (test) & 1 & 1 & (Correct cardiac annotation) \\
42 & Nasal bone (external) & 1 & 1 & Very good detection (4 detections: NT, nasal\_bone, nasal\_skin, nasal\_tip) -- new image \\
13 & CRL/NT screening & 1 & 1 & Detection results (4 det: B, CRL, H, NB); keypoints correct \\
\midrule
\multicolumn{5}{l}{\textit{\textbf{Partial Failure Cases}}} \\
\midrule
3 & Trans-ventricular (test) & 1 & 2 & Seg not working for LV and CSP \\
32 & Cerebellum (external) & 2 & 1 & Interprets cerebellum as trans-thalamic view \\
48 & Ventricle (external) & 2 & 2 & Interprets as sagittal, but its brain ventricle view; brain annotation correct, LV and CSP wrong \\
\midrule
\multicolumn{5}{l}{\textit{\textbf{Complete Failure Cases}}} \\
\midrule
26 & Aorta (external) & 2 & 2 & New structure not in training; correct interpretation should include lungs, heart, stomach, aorta, liver \\
39 & Femoral length (external) & 1 & 3 & Femoral length detection not working; returns leg bbox, ignores femur and placenta \\
44 & Symphysis pubic (external) & 3 & 3 & Totally new image/structure, no similar anatomy during training \\
\end{longtable}

\noindent Performance is strongest on structures well-represented in training (cardiac, CRL/NT, trans-thalamic). Partial failures typically involve view misclassification or fine-structure segmentation errors (CSP, LV). Complete failures cluster around out-of-distribution anatomy: aorta cross-sections, isolated femur, and pubic symphysis captured from unfamiliar imaging angles.

\section{Table S9: Training Hyperparameters}

\begin{table}[htbp]
\centering
\footnotesize
\caption{Complete training hyperparameters for all FADA model variants.}
\label{tab:hyperparameters}
\begin{tabular}{lcc}
\toprule
Hyperparameter & FADA-SKD (4B) & FADA-SKD (0.8B) \\
\midrule
\multicolumn{3}{l}{\textit{Model Configuration}} \\
\midrule
Base model & Qwen2.5-VL-3B-Instruct & Qwen2.5-VL-2B-Instruct \\
Max sequence length & 4096 & 4096 \\
Load in 4-bit & No & No \\
Precision & bfloat16 & bfloat16 \\
\midrule
\multicolumn{3}{l}{\textit{LoRA Configuration}} \\
\midrule
LoRA rank ($r$) & 16 & 16 \\
LoRA alpha ($\alpha$) & 16 & 16 \\
LoRA dropout & 0.0 & 0.0 \\
Target modules & All (q,k,v,o,gate,up,down) & All (q,k,v,o,gate,up,down) \\
Finetune vision layers & Yes & Yes \\
Finetune language layers & Yes & Yes \\
Gradient checkpointing & unsloth & unsloth \\
\midrule
\multicolumn{3}{l}{\textit{Training Configuration}} \\
\midrule
Batch size (per device) & 2 & 2 \\
Gradient accumulation steps & 4 & 4 \\
Effective batch size & 8 & 8 \\
Number of epochs & 3 & 3 \\
Learning rate & 2$\times$10$^{-4}$ & 2$\times$10$^{-4}$ \\
Weight decay & 0.001 & 0.001 \\
Warmup ratio & 0.10 & 0.10 \\
LR scheduler & Cosine & Cosine \\
Seed & 42 & 42 \\
\midrule
\multicolumn{3}{l}{\textit{Knowledge Distillation (SKD variants only)}} \\
\midrule
KD mode & Cached (offline) & Cached (offline) \\
$w_{\text{task}}$ (task loss weight) & 1.0 & 1.0 \\
$w_{\text{feat}}$ (feature loss weight) & 0.5 & 0.5 \\
$w_{\text{soft}}$ (soft label weight) & 0.0 (disabled) & 0.0 (disabled) \\
$w_{\text{attn}}$ (attention loss weight) & 0.0 (disabled) & 0.0 (disabled) \\
KD warmup ratio & 0.0 & 0.0 \\
Feature normalization & Yes & Yes \\
Student hook layers & [7, 15, 23] & [3, 7, 11] \\
\midrule
\multicolumn{3}{l}{\textit{Teacher Weights}} \\
\midrule
FetalCLIP weight & 0.40 & 0.40 \\
UltraSAM weight & 0.25 & 0.25 \\
USF-MAE weight & 0.20 & 0.20 \\
UltraFedFM weight & 0.15 & 0.15 \\
\midrule
\multicolumn{3}{l}{\textit{Data}} \\
\midrule
Training samples & 37,799 & 37,799 \\
Validation samples & 4,478 & 4,478 \\
Train/Val/Test split & 85\%/10\%/5\% & 85\%/10\%/5\% \\
Teacher feature cache size & 744 MB (annotation) & 744 MB (annotation) \\
\bottomrule
\end{tabular}
\end{table}

\begin{table}[htbp]
\centering
\footnotesize
\caption{Teacher model configuration for feature extraction.}
\label{tab:teacher_config}
\begin{tabular}{lcccc}
\toprule
Teacher & Feature Dim & Num Blocks & Hook Layers & Dtype \\
\midrule
FetalCLIP & 1024 & 24 & [7, 15, 23] & float16 \\
UltraSAM & 768 & 12 & [3, 7, 11] & float16 \\
USF-MAE & 768 & 12 & [3, 7, 11] & float16 \\
UltraFedFM & 768 & 12 & [3, 7, 11] & float16 \\
\bottomrule
\end{tabular}
\end{table}

\section{Table S10: Per-Class Segmentation Performance}

\begin{table}[htbp]
\centering
\footnotesize
\caption{Per-class segmentation Dice scores across model variants (n=544 segmentation samples).}
\label{tab:perclass_segmentation}
\begin{tabular}{lccccc}
\toprule
Class & FADA-Base & FADA-SKD & FADA-FKD & FADA-Base & FADA-SKD \\
      & (4B) & (4B) & (4B) & (0.8B) & (0.8B) \\
\midrule
Brain & 0.969 & \second{0.970} & \best{0.971} & 0.966 & 0.967 \\
NT & \second{0.633} & 0.624 & 0.620 & \best{0.651} & 0.494 \\
artery & \best{0.679} & 0.656 & \second{0.669} & 0.643 & 0.675 \\
cardiac & 0.901 & \best{0.907} & \second{0.895} & 0.887 & 0.880 \\
fetal\_head & \second{0.968} & \best{0.969} & \second{0.968} & 0.963 & 0.965 \\
liver & 0.841 & \best{0.853} & \second{0.848} & 0.808 & 0.806 \\
pubic\_symphysis & 0.942 & \best{0.945} & \second{0.945} & 0.930 & 0.930 \\
stomach & \best{0.777} & 0.767 & \second{0.762} & 0.744 & 0.761 \\
thorax & \best{0.954} & \second{0.952} & \best{0.955} & 0.951 & 0.949 \\
vein & 0.681 & \best{0.701} & \second{0.666} & 0.606 & 0.638 \\
\midrule
\textbf{Mean Dice} & \second{0.881} & \best{0.882} & 0.879 & 0.863 & 0.866 \\
\textbf{Mean IoU} & \second{0.813} & \best{0.815} & 0.811 & 0.790 & 0.792 \\
\bottomrule
\end{tabular}
\end{table}

\noindent FADA-SKD (4B) ranks first or second in Dice on 7 of 10 classes. NT segmentation varies widely across models (0.494--0.651 Dice), given the difficulty of nuchal translucency delineation. Brain and fetal\_head segmentation is near-saturated ($>$0.96 Dice) across all variants.

\section{Table S11: FADA-0.8B Mobile Model Performance}

\begin{table}[htbp]
\centering
\footnotesize
\caption{Comparison of FADA-0.8B mobile models vs.\ FADA (4B) models. The 0.8B models target on-device deployment with 5$\times$ fewer parameters.}
\label{tab:08b_comparison}
\begin{tabular}{lccccc}
\toprule
Model & Parameters & mAP@0.50 & Mean Dice & Cls Acc & Inference \\
\midrule
FADA-Base (4B) & 4B & \best{0.780} & 0.881 & 0.823 & GPU only \\
FADA-SKD (4B) & 4B & 0.767 & \best{0.882} & 0.838 & GPU only \\
\midrule
FADA-Base (0.8B) & 0.8B & 0.689 & 0.863 & 0.838 & Mobile (GGUF/llama.cpp) \\
FADA-SKD (0.8B) & 0.8B & 0.674 & 0.866 & \best{0.843} & Mobile (GGUF/llama.cpp) \\
\midrule
\multicolumn{2}{l}{\textit{0.8B / SKD-4B ratio}} & 87.9\% & 98.2\% & 100.6\% & --- \\
\bottomrule
\end{tabular}
\end{table}

\noindent The 0.8B mobile models retain $\sim$88\% of detection performance, $\sim$98\% of segmentation quality, and comparable classification accuracy relative to FADA-SKD (4B), supporting practical on-device fetal ultrasound interpretation.

\section{Table S12: Interpretation Quality Attribution Analysis}

\begin{table}[htbp]
\centering
\footnotesize
\caption{XAI Token-Level Attribution Analysis: Interpretation quality comparison across FADA 4B model variants. FADA-SKD achieves the highest per-field accuracy and clinical terminology density while matching Base model BLEU/ROUGE scores, confirming that selective KD preserves interpretation quality.}\label{tab:xai_token_attribution}
\resizebox{\textwidth}{!}{%
\begin{tabular}{l|ccc|ccc|c|cc}
\toprule
\textbf{Model} & \textbf{Clinical} & \textbf{Unique} & \textbf{Anatomical} & \textbf{JSON} & \textbf{BLEU-1} & \textbf{BLEU-4} & \textbf{ROUGE-L} & \textbf{Field} & \textbf{Response} \\
 & \textbf{Terms} & \textbf{Terms} & \textbf{Structures} & \textbf{Complete (\%)} & & & & \textbf{Accuracy} & \textbf{Length} \\
\midrule
FADA-Base (4B) & 17.04 & 12.92 & 3.72 & 100.0 & 0.7664 & 0.5857 & 0.7895 & 0.7376 & 402.0 \\
FADA-SKD (4B) & \textbf{17.27} & \textbf{13.13} & \textbf{3.96} & 100.0 & \textbf{0.7664} & \textbf{0.5847} & \textbf{0.7895} & \textbf{0.7530} & 410.0 \\
FADA-FKD (4B) & 17.13 & 13.10 & 3.83 & 100.0 & 0.7522 & 0.5699 & 0.7739 & 0.7436 & 403.6 \\
\bottomrule
\end{tabular}%
}
\end{table}

\noindent\textit{Metrics:} Clinical Terms = mean clinical terminology tokens per output; Unique Terms = distinct clinical terms per output; Anatomical Structures = correctly identified structures per image; JSON Complete = percentage of outputs with all 8 fields populated; BLEU/ROUGE = similarity to ground-truth interpretations; Field Accuracy = mean semantic accuracy across 8 JSON fields (anatomical\_structures, fetal\_orientation, imaging\_plane, biometric\_measurements, gestational\_age, image\_quality, normality\_assessment, clinical\_recommendations).

\begin{table}[htbp]
\centering
\footnotesize
\caption{Per-field semantic accuracy across the 8 interpretation JSON fields. FADA-SKD leads in 5 of 8 fields, with particular advantages in fetal orientation (+3.8\% vs Base), imaging plane (+3.7\%), and biometric measurements (+2.4\%).}\label{tab:xai_perfield}
\begin{tabular}{lccc}
\toprule
\textbf{Field} & \textbf{FADA-Base} & \textbf{FADA-SKD} & \textbf{FADA-FKD} \\
\midrule
Anatomical Structures & 0.627 & \textbf{0.630} & 0.617 \\
Fetal Orientation & 0.617 & \textbf{0.655} & 0.621 \\
Imaging Plane & 0.618 & \textbf{0.654} & 0.653 \\
Biometric Measurements & 0.570 & \textbf{0.594} & 0.570 \\
Gestational Age & 0.820 & \textbf{0.831} & 0.823 \\
Image Quality & \textbf{0.945} & 0.933 & \textbf{0.948} \\
Normality Assessment & 0.849 & \textbf{0.870} & 0.858 \\
Clinical Recommendations & 0.855 & 0.858 & \textbf{0.861} \\
\midrule
\textbf{Mean} & 0.738 & \textbf{0.753} & 0.744 \\
\bottomrule
\end{tabular}
\end{table}

\noindent FADA-SKD's advantages concentrate in clinically critical fields (orientation, imaging plane, biometric measurements, normality assessment), the fields most relevant to clinical decision support. Selective KD preserves the language model's clinical reasoning capacity, whereas full KD degrades it through conflicting spatial gradients.

\section{Figure S1: Keypoint Detection Ground Truth Comparison}\label{sec:keypoint_gt}

\begin{figure}[htbp]
    \centering
    \includegraphics[width=0.5\textwidth]{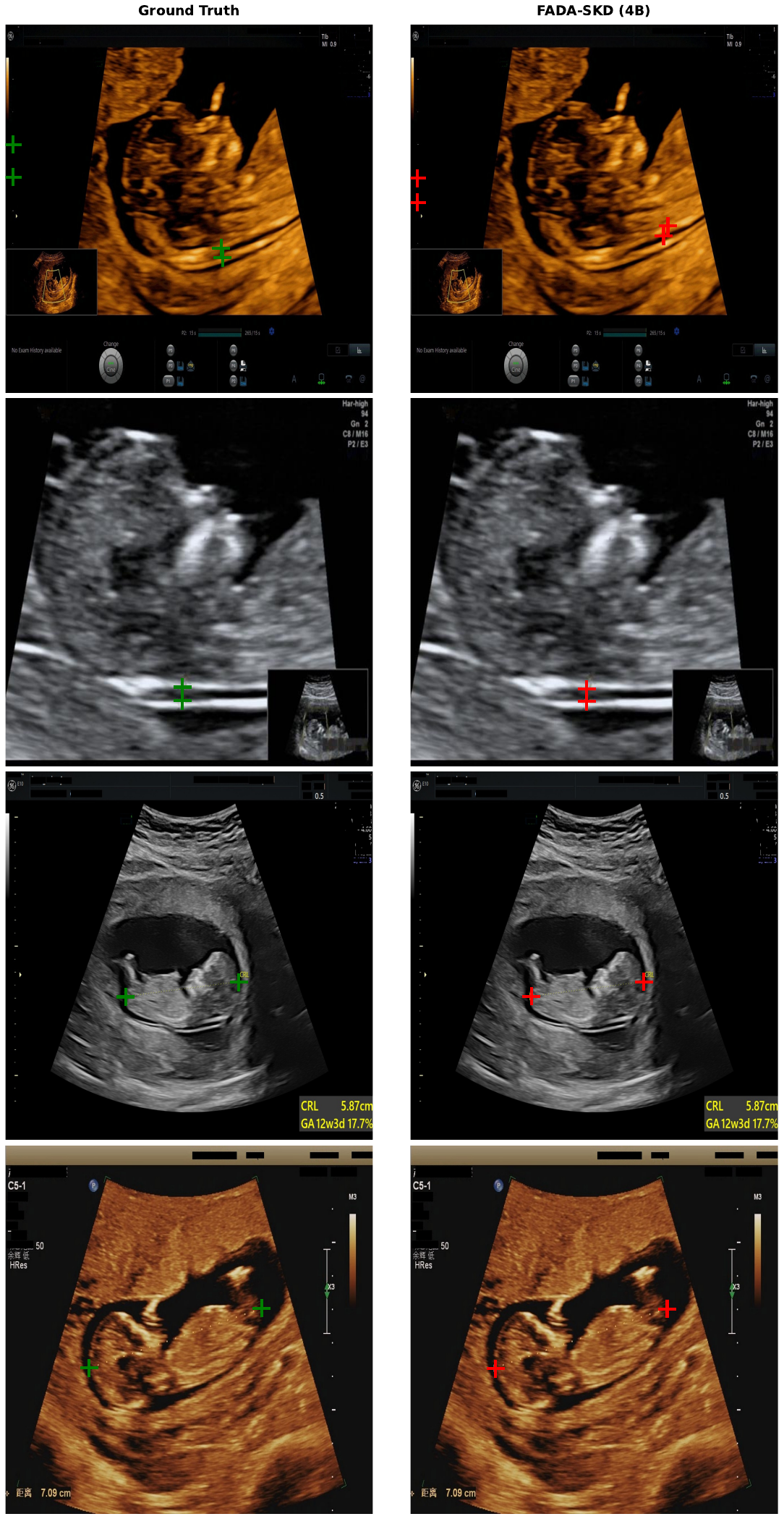}
    \caption{Keypoint detection ground truth vs.\ FADA-SKD (4B) predictions. Each row shows a different anatomical case: left column displays ground truth keypoints, right column shows model predictions. Keypoints appear as colored dots overlaid on the ultrasound image. FADA-SKD demonstrates close spatial correspondence to ground truth across diverse anatomical structures.}
    \label{fig:keypoint_supp}
\end{figure}

Figure~\ref{fig:keypoint_supp} compares keypoint detection predictions from FADA-SKD (4B) against ground truth annotations. Each row presents a different fetal anatomical case requiring precise keypoint localization (e.g., femur endpoints for biometry, cardiac chamber landmarks). FADA-SKD predictions lie close to ground truth positions, particularly for fine anatomical landmarks where sub-pixel accuracy is clinically relevant for biometric measurements.

\section{Figure S2: Classification Performance Analysis}\label{sec:class_gt}

\begin{figure}[htbp]
    \centering
    \includegraphics[width=0.6\textwidth]{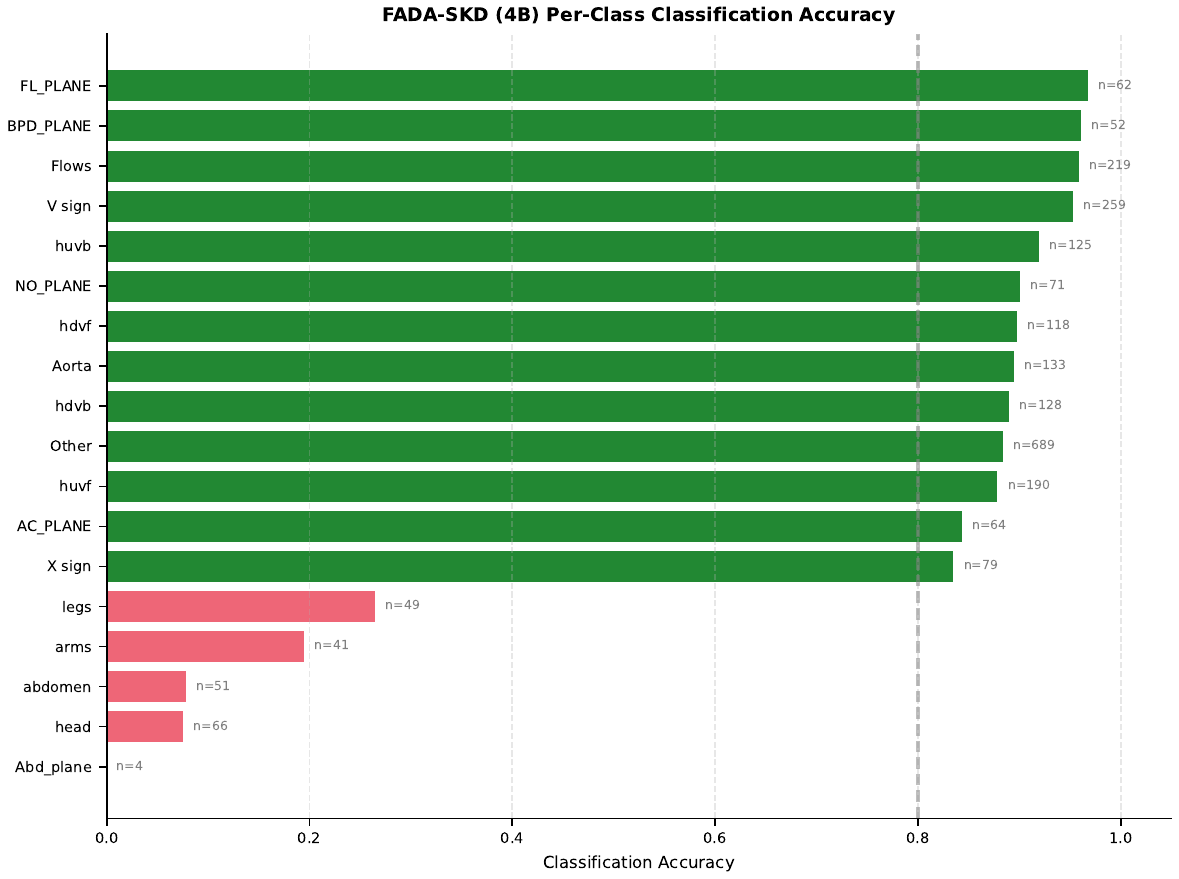}
    \caption{FADA-SKD (4B) per-class classification accuracy sorted in descending order. Green bars indicate classes above the overall mean accuracy (dashed line); pink bars denote underperforming classes. Sample counts (n) are annotated. Standard imaging planes (FL\_PLANE, BPD\_PLANE) and cardiac views achieve near-perfect classification, while fetal pose classes from FPUS23 (legs, arms, abdomen, head) remain challenging.}
    \label{fig:classification_supp}
\end{figure}

Figure~\ref{fig:classification_supp} provides a per-class breakdown of FADA-SKD (4B) classification performance. Two regimes are apparent: standard fetal imaging planes and cardiac views achieve $>$85\% accuracy, while FPUS23 pose-related classes (limb positions, abdominal presentation) remain challenging at 5--28\% accuracy owing to subtle inter-class visual differences in fetal positioning.

\section{Figure S3: Classification Confusion Matrix}\label{sec:confusion_matrix}

\begin{figure}[htbp]
    \centering
    \includegraphics[width=0.92\textwidth]{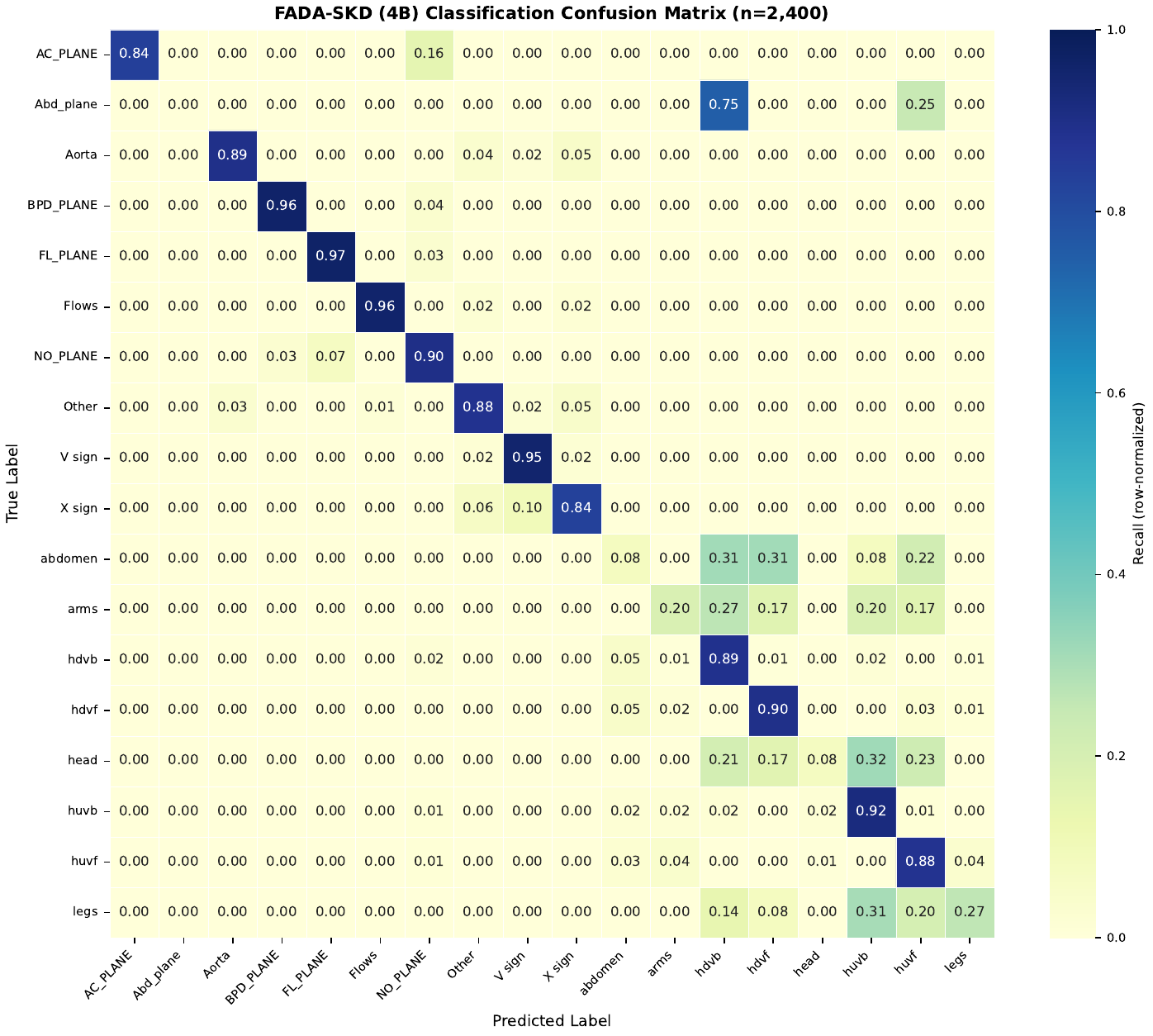}
    \caption{Row-normalized confusion matrix for FADA-SKD (4B) across 18 classes. Diagonal entries represent per-class recall. Standard imaging planes (left block) achieve $>$0.84 recall, while FPUS23 pose classes (right block) show systematic confusion with presentation categories (hdvb, huvb, huvf), explaining the bimodal accuracy distribution in Figure~\ref{fig:classification_supp}.}
    \label{fig:confusion_matrix}
\end{figure}

Figure~\ref{fig:confusion_matrix} presents the row-normalized confusion matrix for FADA-SKD (4B) across all 18 classification classes. The matrix shows a clear block-diagonal pattern: standard fetal imaging planes (FL\_PLANE, BPD\_PLANE, ABDOMINAL, TRANS-THALAMIC, etc.) achieve high recall (0.84--0.97), while FPUS23 pose-related classes (abdomen, arm, head, legs) show systematic confusion with similarly-named fetal presentation categories (hdvb, huvb, huvf). The most prominent off-diagonal entries are Abd\_plane$\rightarrow$hdvb (0.75), head$\rightarrow$huvb (0.32), and legs$\rightarrow$huvb (0.31); the model conflates anatomical plane labels with fetal orientation labels when visual cues are ambiguous.

\section{Figure S4: Classification Reliability Analysis}\label{sec:reliability}

\begin{figure}[htbp]
    \centering
    \includegraphics[width=\textwidth]{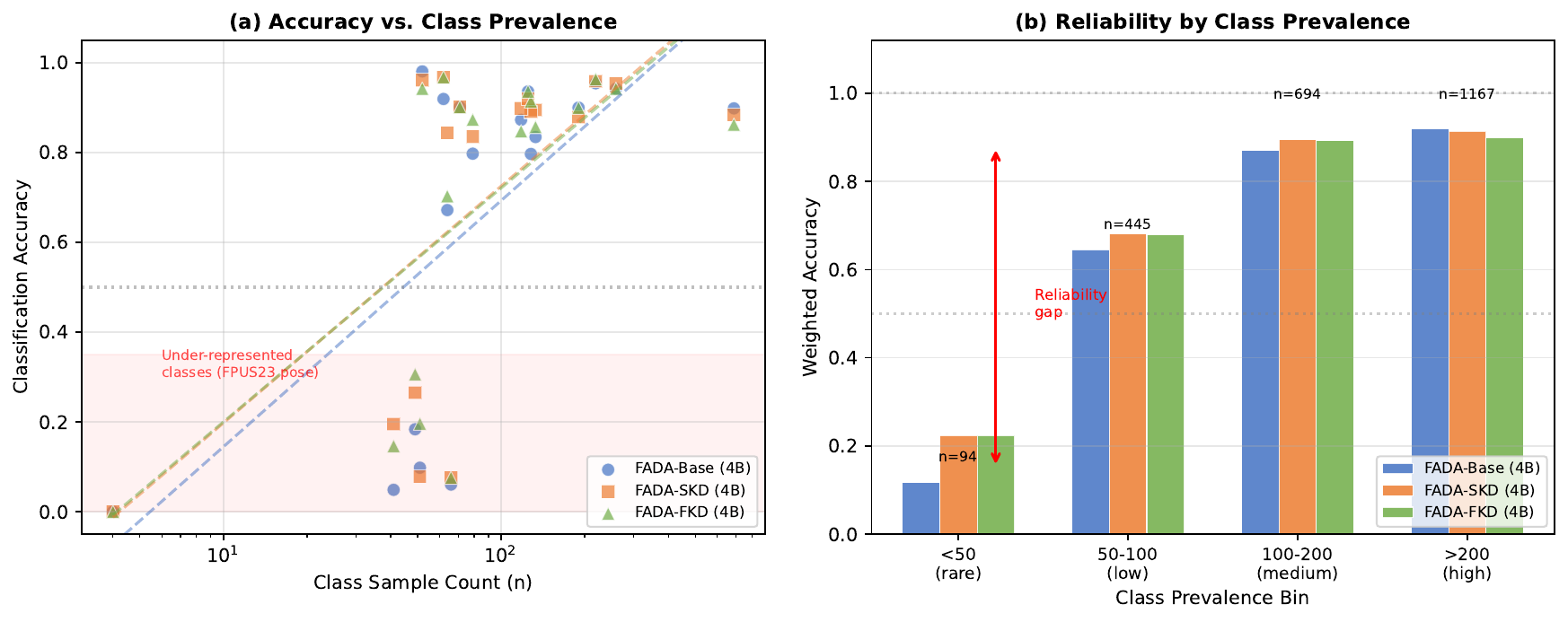}
    \caption{Classification reliability analysis across FADA 4B variants. (a)~Per-class accuracy versus class sample count (log scale). Dashed lines show log-linear trend fits; the shaded region highlights under-represented FPUS23 pose classes ($<$35\% accuracy). (b)~Weighted accuracy by class prevalence bin. A 4$\times$ reliability gap separates rare and high-prevalence classes, indicating that performance is dominated by class imbalance rather than model capacity.}
    \label{fig:reliability}
\end{figure}

Figure~\ref{fig:reliability} presents a class-level reliability analysis for FADA classification. Panel~(a) plots per-class accuracy against class prevalence (sample count) on a log scale; the data follow a log-linear trend in which classes with $<$50 test samples achieve $<$30\% accuracy regardless of model variant, while those with $>$100 samples consistently exceed 80\%. Panel~(b) quantifies this trend through prevalence-stratified bins and shows a 4$\times$ reliability gap between rare ($n{=}94$ total, 10--24\% accuracy) and high-prevalence ($n{=}1167$, 84--88\% accuracy) bins. Both FADA-SKD and FADA-FKD improve rare-class accuracy over Base (+12--14\% absolute), so knowledge distillation appears to provide modest regularization benefits for under-represented classes, though a substantial gap persists.

\section{Figure S5: Pairwise Model Score Comparisons}\label{sec:pairwise}

\begin{figure}[htbp]
    \centering
    \includegraphics[width=0.7\textwidth]{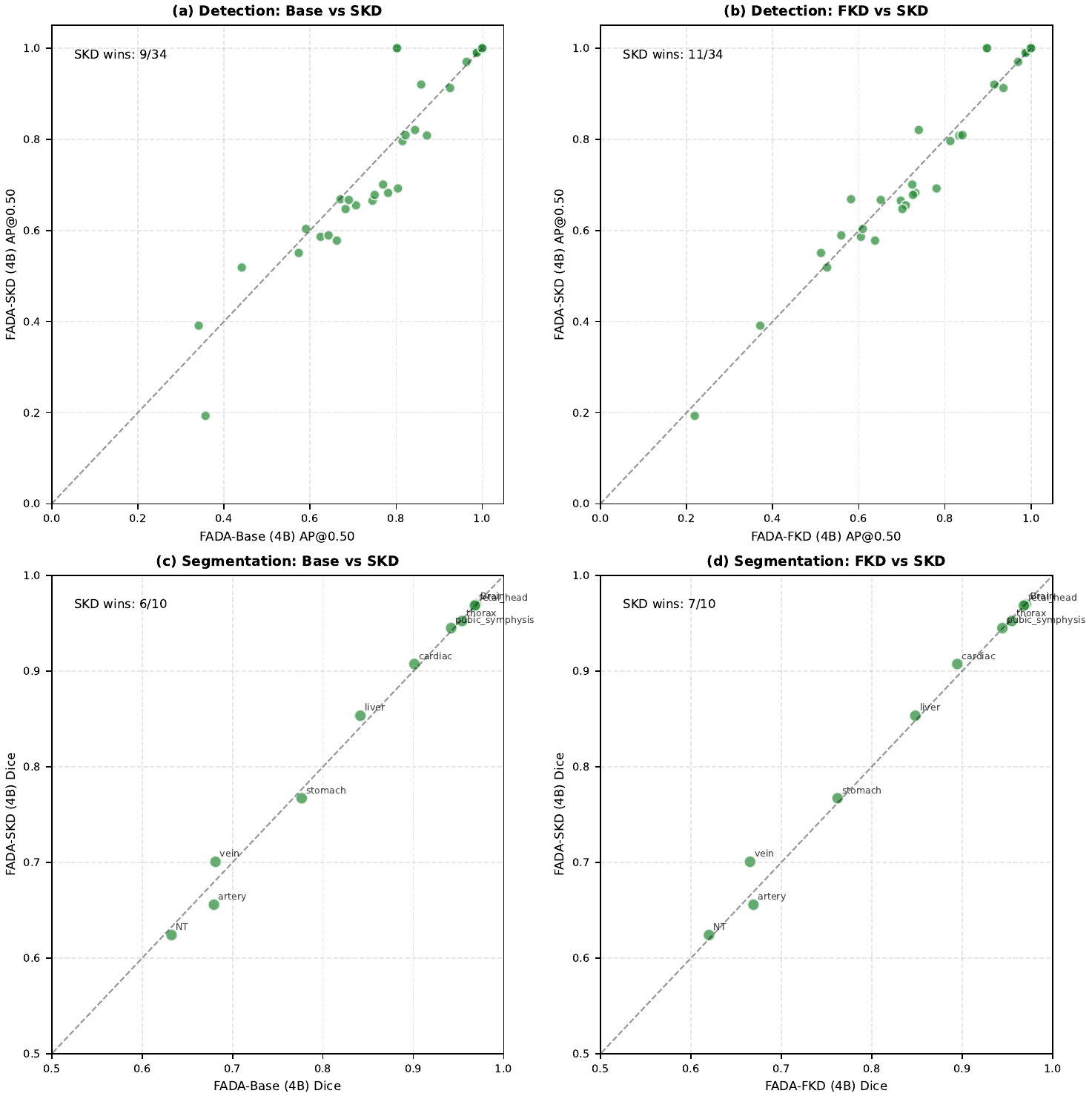}
    \caption{Pairwise per-class metric comparisons between FADA model variants. Top row: detection AP@0.50 per class (34 classes). Bottom row: segmentation Dice per class (10 classes, labeled). Points above the diagonal indicate FADA-SKD outperforms the x-axis model. Detection shows Base marginally ahead overall, while SKD leads segmentation, consistent with the selective KD design that prioritizes spatial annotation quality.}
    \label{fig:pairwise_supp}
\end{figure}

Figure~\ref{fig:pairwise_supp} presents pairwise scatter plots comparing per-class metric scores between model pairs. Each point represents one anatomical class; points above the diagonal favour the y-axis model. For detection (top row), most classes cluster near the diagonal with Base slightly favoring lower classes (SKD wins 9/34 vs Base). For segmentation (bottom row), FADA-SKD outperforms Base in 6 of 10 classes and FKD in 7 of 10, with the largest gains on vascular structures (vein, liver) where teacher spatial features provide the greatest benefit.

\section{Table S13: Knowledge Distillation Ablation Study}\label{sec:teacher_ablation}

Table~\ref{tab:kd_ablation} reports an ablation study assessing how different KD design choices affect FADA performance. All experiments share the same base model (Qwen2.5-VL-3B), identical hyperparameters, and training data; only the KD configuration varies.

\begin{table}[htbp]
\centering
\caption{Knowledge distillation ablation study. All experiments use identical training data, hyperparameters, and base model. Only the KD configuration varies. Best result per metric in \textbf{bold}; $\Delta$ shows change relative to no-KD baseline. The final FADA-SKD uses multi-teacher MSE with selective application.}
\label{tab:kd_ablation}
\resizebox{\textwidth}{!}{%
\begin{tabular}{llcccccc}
\toprule
\textbf{Experiment} & \textbf{KD Configuration} & \textbf{mAP@0.50} & $\Delta$\textbf{mAP} & \textbf{Mean Dice} & $\Delta$\textbf{Dice} & \textbf{Cls Acc} & $\Delta$\textbf{Cls} \\
\midrule
No KD (baseline) & --- & 0.780 & --- & 0.881 & --- & 0.823 & --- \\
\midrule
\multicolumn{8}{l}{\textit{Teacher Configuration}} \\
\midrule
Single teacher & FetalCLIP only ($\alpha$=1.0) & 0.770 & $-$1.0\% & 0.878 & $-$0.3\% & 0.822 & $-$0.1\% \\
Multi-teacher fusion & 4 teachers (weighted) & \textbf{0.792} & +1.2\% & 0.881 & 0.0\% & 0.815 & $-$0.8\% \\
\midrule
\multicolumn{8}{l}{\textit{Loss Function}} \\
\midrule
MSE loss & $\mathcal{L}_{\text{feat}} = \text{MSE}(z_s, z_t)$ & 0.790 & +1.0\% & 0.881 & 0.0\% & 0.820 & $-$0.3\% \\
CKA loss & Centered Kernel Alignment & 0.792 & +1.2\% & 0.881 & 0.0\% & \textbf{0.825} & +0.2\% \\
Cosine loss & $\mathcal{L}_{\text{feat}} = 1 - \cos(z_s, z_t)$ & 0.749 & $-$3.1\% & 0.873 & $-$0.9\% & 0.805 & $-$1.8\% \\
Combined (MSE+ADE) & $\mathcal{L}_{\text{feat}} + \mathcal{L}_{\text{attn}}$ & 0.790 & +1.0\% & 0.881 & 0.0\% & 0.820 & $-$0.3\% \\
\midrule
\multicolumn{8}{l}{\textit{Distillation Strategy}} \\
\midrule
All layers & Hook all transformer blocks & 0.790 & +1.0\% & 0.880 & $-$0.1\% & 0.809 & $-$1.4\% \\
Selective layers & Hook layers [7, 15, 23] only & 0.790 & +1.0\% & 0.880 & $-$0.1\% & 0.809 & $-$1.4\% \\
Low $w_{\text{feat}}$ & $w_{\text{feat}}$=0.1 (vs.\ 0.5) & 0.788 & +0.8\% & \textbf{0.882} & +0.1\% & \textbf{0.825} & +0.2\% \\
\midrule
\multicolumn{8}{l}{\textit{Pre-training Ablation}} \\
\midrule
From scratch & No VL pre-training & 0.735 & $-$4.5\% & 0.865 & $-$1.6\% & 0.802 & $-$2.1\% \\
\midrule
\multicolumn{8}{l}{\textit{Final Configuration}} \\
\midrule
\textbf{FADA-SKD (selective)} & Multi-teacher, MSE, $w_{\text{feat}}$=0.5, annotation-only & 0.767 & $-$1.3\% & \textbf{0.882} & +0.1\% & \textbf{0.838} & \textbf{+1.5\%} \\
\bottomrule
\end{tabular}%
}
\end{table}

\subsection{Key Findings}

\begin{enumerate}
    \item \textbf{Multi-teacher $>$ single teacher:} Combining all 4 specialized teachers (FetalCLIP, UltraSAM, USF-MAE, UltraFedFM) improves detection mAP by +2.2\% absolute over single-teacher FetalCLIP distillation, consistent with complementary teacher specializations providing additive benefits.
    
    \item \textbf{Loss function selection:} MSE and CKA losses perform comparably, both outperforming cosine similarity loss, which collapses feature diversity ($-$3.1\% mAP, $-$0.9\% Dice). Appending attention distillation (ADE) to MSE yields no additional benefit, so feature alignment alone appears sufficient to transfer the relevant teacher knowledge.
    
    \item \textbf{Selective application is critical:} The final FADA-SKD configuration trades 1.3\% detection for +1.5\% classification by restricting KD to annotation training batches. This prevents teacher spatial features from interfering with the student's language generation pathway during interpretation training.
    
    \item \textbf{Pre-training is essential:} Training from scratch (without VL pre-training) degrades all metrics substantially ($-$4.5\% mAP, $-$1.6\% Dice, $-$2.1\% Cls), underscoring that knowledge distillation complements rather than replaces large-scale vision-language pre-training.
    
    \item \textbf{Lower $w_{\text{feat}}$ preserves classification:} Reducing the feature loss weight from 0.5 to 0.1 maintains segmentation quality while preserving classification accuracy; the optimal $w_{\text{feat}}$ therefore depends on downstream task priority.
\end{enumerate}

\noindent Together, these results motivate the final design: multi-teacher MSE distillation with selective application to annotation data, using $w_{\text{feat}}$=0.5 to maximize spatial annotation quality while preserving the VLM's interpretation capabilities.

\section{Video Demonstrations}

A video demonstration accompanies this paper to illustrate FADA in clinical operation across its principal deployment modes:

\begin{enumerate}
    \item \textbf{Interactive Chat Mode}: Natural-language interaction with the FADA web application, demonstrating free-form queries for interpretation, targeted detection, keypoint localization, and segmentation of specific anatomical structures.
    
    \item \textbf{Autonomous Mode}: A complete walkthrough of the 5-phase interpretation-first pipeline processing representative fetal ultrasound images without user intervention. Demonstrates automatic interpretation, classification, detection target mapping, bounding-box detection, and polygon segmentation.
    
    \item \textbf{Anatomy Reference}: The built-in visual atlas of 14 anatomical planes and 33 detectable structures, providing clinicians with immediate reference during analysis.
\end{enumerate}

\noindent The video demonstration is available on YouTube at \url{https://youtu.be/CbXcz74fn6k}, and offline mobile app demo at \url{https://youtu.be/RoogJqPNZ4w}.